\documentclass{article}

\usepackage{iclr2023_conference,times}
\iclrfinalcopy

\usepackage[utf8]{inputenc} %
\usepackage[T1]{fontenc}    %
\usepackage{siunitx}
\DeclareSIUnit{\million}{~\text{M}}
\DeclareSIUnit{\kilo}{~\text{K}}
\usepackage{url}            %
\usepackage{booktabs}       %
\usepackage{amsfonts}       %
\usepackage{nicefrac}       %
\usepackage{microtype}      %
\usepackage{multirow}
\usepackage{amsmath}
\usepackage{amssymb}
\usepackage{amsthm}
\usepackage[disable]{todonotes}
\usepackage{comment}
\usepackage{graphicx,subcaption}
\usepackage[subtle]{savetrees}
\usepackage[font=normalsize,labelfont={bf}]{caption}
\usepackage{parskip}
\usepackage{xr-hyper}
\usepackage{hyperref}
\usepackage{comment}
\hypersetup{
    colorlinks=true,
    linkcolor=black,
    filecolor=magenta,      
    urlcolor=blue,
    citecolor=black
}       %

\usepackage{graphicx}

\newcommand{\bfa}{\mathbf{a}}
\newcommand{\bfu}{\mathbf{u}}
\newcommand{\bfr}{\mathbf{r}}

\newcommand{\bfc}{\mathbf{c}}
\newcommand{\bfx}{\mathbf{x}}
\newcommand{\bfy}{\mathbf{y}}
\newcommand{\bfz}{\mathbf{z}}
\newcommand{\bfe}{\mathbf{e}}

\newcommand{\bflambda}{\boldsymbol{\lambda}}
\newcommand{\bfmu}{\boldsymbol{\mu}}

\newcommand{\tinyk}{\text{\scalebox{0.75}{\hspace{-0.5pt}$k$}}}
\newcommand{\tinykm}{\text{\scalebox{0.75}{\hspace{-0.5pt}$k-1$}}}

\newcommand{\nk}{\ensuremath{{n_\tinyk}}}

\newcommand{\sk}{\ensuremath{{s_\tinyk}}}
\newcommand{\skm}{\ensuremath{{s_\tinykm}}}

\newcommand{\ek}{\ensuremath{{e_\tinyk}}}

\newcommand{\scx}{\textsc{x}}

\newcommand{\bfW}{\ensuremath{\mathbf{W}}}

\newcommand{\bfb}{\ensuremath{\mathbf{b}}}

\newcommand{\dti}{^{\langle t \rangle}}
\newcommand{\dtim}[1]{^{\langle t-{#1} \rangle}}

\newcommand{\hp}{{\,\mathrel{\vcenter{\hbox{\tiny$\odot$}}}\,}}

\newcommand{\bwrap}[1]{\left( #1 \right)}
\newcommand{\concat}[1]{\left[ #1 \right]}
\newcommand{\fun}[2]{{#1}\bwrap{#2}}

\newtheorem{theorem}{Theorem}
\newtheorem{corollary}{Corollary}
\newtheorem{proposition}{Proposition}

\usepackage{filecontents}

\makeatletter
\newcommand*{\addFileDependency}[1]{%
  \typeout{(#1)}
  \@addtofilelist{#1}
  \IfFileExists{#1}{}{\typeout{No file #1.}}
}
\makeatother

\newcommand*{\myexternaldocument}[1]{%
    \externaldocument{#1}%
    \addFileDependency{#1.tex}%
    \addFileDependency{#1.aux}%
}

\myexternaldocument{./supplement}

\title{Efficient recurrent architectures through activity sparsity and sparse back-propagation through time}

\author{Anand Subramoney$^{1,2,\footnotemark[1]{}}$, Khaleelulla Khan Nazeer$^3$, Mark Sch\"one$^3$, Christian Mayr$^{3,4}$, David Kappel$^1$ \\
$^1$ Institute for Neural Computation, Ruhr University Bochum, Germany \\
$^2$ Royal Holloway, University of London \\
$^3$ Faculty of Electrical and Computer Engineering, Technische Universit\"at Dresden, Dresden, Germany \\
$^4$ Centre for Tactile Internet with Human-in-the-Loop (CeTI),Technische Universit\"at Dresden, Dresden, Germany  \\
\texttt{\{anand.subramoney,david.kappel\}@ini.rub.de} \\
\texttt{\{khaleelulla.khan\_nazeer,mark.schoene,christian.mayr\}@tu-dresden.de}\\
}

\begin{document}

\maketitle
\footnotetext[1]{Work done while at Ruhr University Bochum}

\begin{abstract}

Recurrent neural networks (RNNs) are well suited for solving sequence tasks in resource-constrained systems due to their expressivity and  low computational requirements.
However, there is still a need to bridge the gap between what RNNs are capable of in terms of efficiency and performance and real-world application requirements.
The memory and computational requirements arising from propagating the activations of all the neurons at every time step to every connected neuron, together with the sequential dependence of activations, contribute to the inefficiency of training and using RNNs.
We propose a solution inspired by biological neuron dynamics that makes the communication between RNN units sparse and discrete.
This makes the backward pass with backpropagation through time (BPTT) computationally sparse and efficient as well.
We base our model on the gated recurrent unit (GRU), extending it with units that emit discrete events for communication triggered by a threshold so that no information is communicated to other units in the absence of events. 
We show theoretically that the communication between units, and hence the computation required for both the forward and backward passes, scales with the number of events in the network.
Our model achieves efficiency without compromising task performance, demonstrating competitive performance compared to state-of-the-art recurrent network models in real-world tasks, including language modeling.
The dynamic activity sparsity mechanism also makes our model well suited for novel energy-efficient neuromorphic hardware.
Code is available at \url{https://github.com/KhaleelKhan/EvNN/}.

\end{abstract}

\section{Introduction}
\label{sec:introduction}
Large scale models such as GPT-3~\citep{brown_Language_2020} and DALL-E~\citep{ramesh_ZeroShot_2021} have demonstrated that scaling up deep learning models to billions of parameters improve not just their performance but lead to entirely new forms of generalisation.
But for resource constrained environments, transformers are impractical due to their computational and memory requirements during training as well as inference.
Recurrent neural networks (RNNs) may provide a viable alternative in such low-resource environments, but require further algorithmic and computational optimizations.
While it is unknown if scaling up recurrent neural networks can lead to similar forms of generalization, the limitations on scaling them up preclude studying this possibility.
The dependence of each time step's computation on the previous time step's output prevents easy parallelisation of the model computation.
Moreover, propagating the activations of all the units in each time step is computationally inefficient and leads to high memory requirements when training with backpropagation through time (BPTT).

While allowing extraordinary task performance, the biological brain's recurrent architecture is extremely energy efficient \citep{mead2020we}.
One of the brain's strategies to reach these high levels of efficiency is activity sparsity.
In the brain, (asynchronous) event-based and activity-sparse communication results from the properties of the specific physical and biological substrate on which the brain is built.
Biologically realistic spiking neural networks and neuromorphic hardware aim to use these principles to build energy-efficient software and hardware models \citep{roy2019towards,schuman2017survey}. 
However, despite progress in recent years, their task performance has been relatively limited for real-world tasks compared to recurrent architectures based on LSTM and GRU.

In this work, we propose an activity sparsity mechanism inspired by biological neuron models, to reduce the computation required by RNNs at each time step.
Our method adds a mechanism to the recurrent units to emit discrete events for communication triggered by a threshold so that no information is communicated to other units in the absence of events. 
With event-based communication, units in the model can decide when to send updates to other units, which then trigger the update of receiving units. 
When events are sent sparingly, this leads to activity-sparsity where most units do not send updates to other units most of the time, leading to substantial computational savings during training and inference.
We formulate the gradient updates of the network to be sparse using a novel method, extending the benefit of the computational savings to training time.
We theoretically show, in the continuous time limit, that the time complexity of calculating weight updates is proportional to the number of events in the network.
We demonstrate these properties using Gated Recurrent Unit (GRU)~\citep{cho2014learning} as a case study, and call our model Event-based Gated Recurrent Unit (EGRU).
We note, however, that our dynamic activity-sparsity mechanism can be applied to any RNN architecture.

In summary, the main contributions of this paper are the following:
\begin{enumerate}
    \item We introduce a variant of the GRU  with an event-generating mechanism, called the EGRU.
    \item We theoretically show that, in the continuous time limit, both the forward pass computation and the computation of parameter updates in the EGRU scales with the number of events (active units).
    \item We demonstrate that the EGRU exhibits task-performance competitive with state-of-the-art recurrent network architectures on real-world machine learning benchmarks.
    \item We empirically show that EGRU exhibits high levels of activity-sparsity during both inference (forward pass) and learning (backward pass).
\end{enumerate}
We note here that methods for training with parameter sparsity or improving handling of long-term dependencies are both orthogonal to, and can be combined with our approach (which we plan to do in future work).
Our focus, in this paper, is exclusively on using activity-sparsity to increase the efficiency of RNNs, specifically the GRU.
We expect our method to be more efficient but not better at handling long-range dependencies compared to the GRU.

The sparsity of the backward-pass overcomes one of the major roadblocks in using large recurrent models, which is having enough computational resources to train them.
We demonstrate the task performance and activity sparsity of the model implemented in PyTorch, but this formulation will also allow the model to run efficiently on CPU-based nodes when implemented using appropriate software paradigms.
Moreover, an implementation on novel neuromorphic hardware like~\citet{davies2018loihi,hoppner2017dynamic}, that is geared towards event-based computation, can make the model orders of magnitude more energy efficient~\citep{ostraubenchmarking}.

\section{Related work}
Activity sparsity in RNNs has been proposed previously~\citep{neil_Delta_2017, neil_Phased_2016,hunter2021sparsities}, but only focusing on achieving it during inference.
Conditional computation is a form of activity sparsity used in~\citet{fedus_Switch_2021} to scale a feedforward transformer architecture to 1 trillion parameters.
An asynchronous event-based architecture was recently proposed specifically targeted towards graph neural networks~\citep{schaefer2022aegnn}.
QRNNs~\citep{bradbury_QuasiRecurrent_2016}, SRUs~\citep{lei_Simple_2017} and IndRNNs~\citep{li_Independently_2018} target increasing the parallelism in a recurrent network without directly using activity-sparsity.
Unlike~\citet{fedus_Switch_2021} that used a separate network to decide which sub-networks should be active~\citep{shazeer_Outrageously_2017}, our architecture uses a unit-local decision making process for the dynamic activity-sparsity.
The cost of computation is lower in our model compared to~\citet{neil_Delta_2017}, and can be implemented to have parallel computation of intermediate updates between events, while also being activity sparse in its output.

Models based on sparse communication~\citep{yan2022distributed} for scalability have been proposed recently for feedforward networks, using locality sensitivity hashing to dynamically choose downstream units for communicating activations.
This is a dynamic form of parameter-sparsity~\citep{hoefler2021sparsity}, which is orthogonal to and complementary with our method for activity-sparsity, and can be combined for additional gains.
The use of rectified linear units (ReLU) in recurrent networks as in the IRNN~\citep{le2015simple} is closely related, and can lead to sparse activations.
But current literature lacks an analysis of such models from the context of efficiency.
Moreover, the use of ReLU activation requires careful weight initialization, and it seems to lag behind in task performance in comparisons with other RNN models~\citep{li_Independently_2018}.

Biologically realistic spiking networks~\citep{maass1997networks} are often implemented using event-based updates and have been scaled to huge sizes~\citep{jordan_extremely_2018}, albeit without any task-related performance evaluation.
Models for deep learning with recurrent spiking networks~\citep{bellec_Long_2018,salaj_Spike_2021} mostly focus on modeling biologically realistic memory and learning mechanisms.
Moreover, units in a spiking neural network implement dynamics based on biology and communicate solely through unitary events, while units in an EGRU send real-valued signals to other units, and have different dynamics.
A sparse learning rule was recentl y proposed~\citep{bellec_solution_2020} that is a local approximation of backpropagation through time, but not event-based.

The theoretical analysis of the event-based learning rule for the continuous time EGRU is inspired by, and a generalization of the analysis in~\citet{wunderlich_Eventbased_2021} for spiking neurons.
As in that paper, we use the adjoint method for ordinary differential equations (ODEs)~\cite{pontryagin1962mathematical,chen_Neural_2018} combined with sensitivity analysis for hybrid discrete/continuous systems~\citep{galan_Parametric_1999,chen2020learning}.
Using surrogate gradients for backpropagating through the non-differential threshold function was originally proposed for feedforward spiking networks in~\citet{esser_Convolutional_2016} and developed further in~\citet{bellec_Long_2018,zenke_SuperSpike_2018}. 
The sparsity of learning with BPTT when using appropriate surrogate gradients in a discrete-time feed-forward spiking neural network was recently described in~\citet{perez-nieves_Sparse_2021}.

A continuous time version of sigmoidal RNNs was proposed in~\citet{beer_Dynamics_1995} and for GRUs in~\citet{debrouwer2019gruodebayesa}.
The latter used a Bayesian update for network states when input events were received, but the network itself was not event-based.
As in~\citet{neil_Phased_2016,lechner2020learning}, the focus there was on modeling irregularly spaced input data, and not on event-based network simulation or activity-sparse inference and training.
\citet{chang_AntisymmetricRNN_2019} also recently proposed a continuous time recurrent network for more stable learning, without event-based mechanics.
GRUs were formulated in continuous time for analyzing its autonomous dynamics in~\citet{jordan_Gated_2019}.

\section{Event-based GRU}
\label{sec:EGRU}
\subsection{Time-sparse GRU formulation}

\begin{figure}[htbp]
    \centering
    \includegraphics[width=\textwidth]{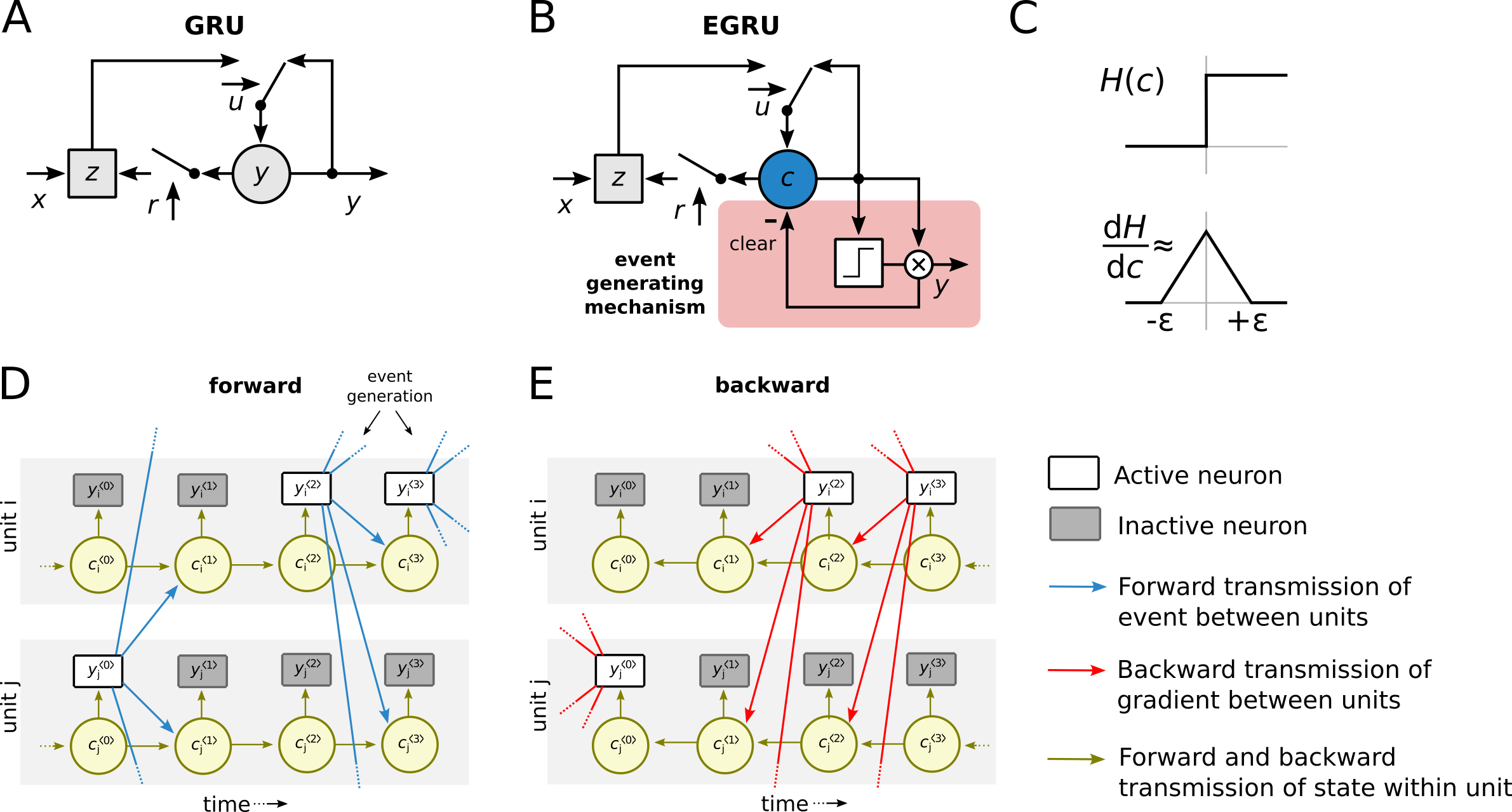}
    \caption{Illustration of EGRU. 
    \textbf{A:} A single unit of the original GRU model adapted from \cite{cho2014learning}. 
    \textbf{B:} EGRU unit with event generating mechanism. 
    \textbf{C:} Heaviside function and surrogate gradient.
    \textbf{D:} Forward state dynamics for two EGRU units ($i$ and $j$).
    \textbf{E:} Activity-sparse backward dynamics for two EGRU units ($i$ and $j$). 
    Note that we only have to backpropagate through units that were active or whose state was close to the threshold at each time step.
    }
    \label{fig:egru-overview}

\end{figure}

We base our model on the GRU~\citep{cho2014learning}, illustrated for convenience in Fig.~\ref{fig:egru-overview}A. 
It consists of internal gating variables for updates ($\bfu$) and a reset ($\bfr$), that determine the behavior of the internal state $\bfy$.
The state variable $\bfz$ determines the interaction between external input $\bfx$ and the internal state. 
The dynamics of a layer of GRU units, at time step $t$, is given by the set of vector-valued update equations:
\begin{equation}
\label{eq:gru-dt}
\begin{split}
     \bfu\dti = \fun{\sigma}{ \bfW_{u}  \concat{ \bfx\dti,\; \bfy\dtim1 } + \bfb_u }\,, & \quad
     \bfr\dti = \fun{\sigma}{\bfW_{r} \concat{ \bfx\dti,\; \bfy\dtim1 } + \bfb_r }\,, \\[2mm]
     \bfz\dti = \fun{g}{ \bfW_z \concat{ \bfx\dti,\;  \bfr\dti \hp\; \bfy\dtim1 } + \bfb_z }\,, & \quad
     \bfy\dti = \bfu\dti \hp\; \bfz\dti  + (1-\bfu\dti) \hp\; \bfy\dtim1\,,
\end{split}
\end{equation}
where $\bfW_{u/r/z}$, $\bfb_{u/r/z}$ denote network weights and biases,  $\hp$ denotes the element-wise (Hadamard) product, and $\fun{\sigma}{\cdot}$ is the vectorized sigmoid function. 
The notation $\concat{ \bfx\dti, \bfy\dtim1 }$ denotes vector concatenation. 
The function $\fun{g}{\cdot}$ is an element-wise nonlinearity, typically the hyperbolic tangent.

We introduce an event generating mechanisms by augmenting the GRU with a rectifier (a thresholding function).
See Fig.~\ref{fig:egru-overview}B for an illustration. 
With this addition the internal state variable $y_i\dti$ is nonzero only when the internal dynamics reach a threshold $\vartheta_i$ and is  cleared immediately afterwards, thus making $y_i\dti$ event-based.
Formally, we add an auxiliary internal state $c_i\dti$ to the model, and replace $\bfy\dti = (y_1\dti,y_2\dti,\dots)$ with the event-based form
\begin{align}
y_i\dti \;=\; c_i\dti \, \fun{H}{ c_i\dti - \vartheta_i } \quad \text{with} \quad
c_i\dti \;=\; u_i\dti z_i\dti  \,+\, (1-u_i\dti) c_i\dtim1 \,-\, y_i\dtim1 \;,
\label{eq:event-mechanism}
\end{align}
where $\fun{H}{ \cdot }$ is the Heaviside step function and $\vartheta_i>0$ is a trainable threshold parameter. 
$\fun{H}{\cdot}$ is the threshold gating mechanism here, generating a single non-zero output when $c_i\dti$ crosses the threshold $\vartheta_i$.
That is, at all time steps $t$ with $c_i\dti < \vartheta_i, \forall i$,  we have $y_i\dti = 0$. 
The $-\, y_i\dtim1$ term in Eq.~\eqref{eq:event-mechanism} makes emission of multiple consecutive events by the same unit  unlikely, hence favoring overall sparse activity.
With this formulation, each unit only needs to be updated when an input is received either externally or from another unit in the network.
This is because, if both $x_i\dti = y_i\dtim1 = 0$ for the $i$-th unit, then $u_i\dti,\; r_i\dti,\; z_i\dti$ are essentially constants, and hence the update for $y_i\dti$ can be retroactively calculated efficiently on the next incoming event.

\subsection{Sparse approximate BPTT}\label{sec:discrete-learning}

The threshold activation function $H(c)$ to decide whether to emit an event in Eq.~\eqref{eq:event-mechanism} is not differentiable at the threshold $\vartheta_i$. 
We define a surrogate gradient at that point for calculating the backpropagated gradients.
The surrogate gradient is defined as a piece-wise linear function that is non-zero for values of state $c_i$ between $\vartheta_i + \varepsilon$ and $\vartheta_i - \varepsilon$ as shown in the inset in Fig.~\ref{fig:egru-overview}C.
Since the surrogate gradient is zero whenever the internal state of the unit is below $\vartheta_i - \varepsilon$, the backpropagated gradients are also $0$ for all such units, making the backward-pass sparse (see Fig.~\ref{fig:egru-overview}D, E for an illustration).
Note that the case where the internal unit state is above $\vartheta_i + \varepsilon$ tends to occur less often, since the unit will emit an event and the internal state will be cleared at the next simulation step in that case.

\subsection{Computation and memory reduction due to sparsity}
\label{sec:sparse-computation}

For the forward pass of the  EGRU, an activity sparsity of $\alpha$ (i.e. an average of $\alpha$ events per simulation step) leads to the reduction of multiply-accumulate operations (MAC), by factor $\alpha$.
We focus on MAC operations, since they are by far the most expensive compute operation in these models.
MACs are also relevant for our focus on resource constrained systems which are more likely to be based on CPUs rather than GPUs.
If optimally implemented an activity sparsity of 80\% will require 80\% fewer MAC operations compared to a standard GRU.
Computation related to external input is only performed at input times, and hence is as sparse as the input, both in time and space.
During the backward pass, a similar factor of computational reduction is observed, based on the backward-pass sparsity $\beta$ which is, in general, less than $\alpha$.
This is because, when the internal state value is not within $\pm \varepsilon$ of the threshold $\vartheta$, the backward pass is skipped, as described in section~\ref{sec:discrete-learning}.
Since our backward pass is also sparse, we need to store only $\beta$ fraction of the activations for later use, hence also reducing the memory usage.
In all our experiments, we report activity-sparsity values calculated through simulations.

\section{Theoretical analysis of the EGRU}
\label{sec:theoritical-analysis}
To further analyze the dynamics of EGRU we develop a continuous-time version of the model. This allows us to study the model parameters as a dynamical system. 

\subsection{Limit to continuous time}

Eq.~\eqref{eq:gru-dt} of the discrete time model  considers the GRU dynamics only at integer time points, $t_0=0,\; t_1=1,\; t_2=2, \dots$.
However, in general it is possible to express the GRU dynamics for an arbitrary time step $\Delta t$, with $t_n = t_{n-1} + \Delta t$.
The discrete time GRU dynamics can be intuitively interpreted as an Euler discretization of an ordinary differential equation (ODE)~\citep{jordan_Gated_2019} (see Supplement), which we extend further to formulate the EGRU.
This is equivalent to taking the continuous time limit $\Delta t \rightarrow 0$ to get dynamics for the internal state $\bfc(t)$.
In the resulting dynamical system equations inputs cause changes to the states only at the event times, whereas the dynamics between events can be expressed through ODEs. 
To arrive at the continuous time formulation we introduce the neuronal activations $\bfa_u(t)$, $\bfa_r(t)$ and $\bfa_z(t)$, with
\begin{align}
\begin{gathered}
  \bfu(t) \;=\; \fun{\sigma}{ \bfa_u(t) }\,, \quad
 \bfr(t) \;=\; \fun{\sigma}{ \bfa_r(t) }\,, \quad
 \bfz(t) \;=\; \fun{g}{ \bfa_z(t) }\,, \\[3mm]
 \text{with dynamics}\quad
 \tau_{s}\, \dot\bfa_\scx \,=\, -\bfa_\scx\ - \bfb_\scx\,, \quad \scx \in \{u,\; r,\; z\} \qquad
\end{gathered}
\label{eq:current-dynamics}
\end{align}
\begin{align}
    \text{and}\quad \tau_{m}\, \dot{\bfc}(t)  \;=\;  \bfu(t) \hp \left( \bfz(t)  - \bfc(t) \right)
    \;=\; \fun{F}{t, \bfa_u, \bfa_r, \bfa_z,\bfc}
    \;,
    \label{eq:unperturbed-dynamics}
\end{align}
where $\tau_{s}$ and $\tau_{m}$ are time constants, $\bfc(t)$, $\bfu(t)$ and $\bfz(t)$ are the continuous time analogues to $\bfc\dti$, $\bfu\dti$ and $\bfz\dti$, and $\dot\bfa_\scx$ denotes the time derivative of $\bfa_\scx$.
The boundary conditions are defined for $t=0$ as $\bfa_\scx(0)=\bfc(0)=\mathbf{0}$.
The function $F$ in Eq.~\eqref{eq:unperturbed-dynamics} determines the behavior of the EGRU between event times, i.e. when $\bfx(t)=\mathbf{0}$ and $\bfy(t)=\mathbf{0}$. 
Nonzero external inputs and internal events cause jumps in $\bfc(t)$ and $\bfa_\scx(t)$.
For theoretical tractability, we add a decay term $-\bfa_\scx$ to the ODE in Eq.~\eqref{eq:current-dynamics}, which is implemented with a small or zero time constant in the discrete time model.

To describe these dynamics we introduce the set of internal events $\bfe$, $\ek \in \bfe$, $\ek = \bwrap{\sk,\nk}$, where $\sk$ are the continuous (real-valued) event times, and $\nk$ denotes which unit got activated.
The formulation of the event generating mechanisms Eq.~\eqref{eq:event-mechanism} introduced above can be expressed as event $\ek$ that is triggered whenever $c_{\nk}(t)$ reaches $\vartheta$. More precisely:
\begin{align}
    \bwrap{\sk,\nk} \;:\; c_{\nk}^-(\sk) = \vartheta_{\nk}\;,
\end{align}
where the superscript $.^-$ ($.^+$) denotes the quantity just before (after) the event.
The clearing mechanism in continuous time is expressed as resetting $c_i(s)$ to zero right after event times $s$. 
This is because in continuous time the exact time $s$ at which the internal variable $c_i(s)$ reaches the threshold ($c_i(s) = \vartheta_i$) can be determined with very high precision. 
Therefore, the value of $c_i(s)$ and the instantaneous amplitude of $y_i(s)$ simultaneously approach $\vartheta_i$ at time point $s$, so that the $-y_i$ term in Eq.~\eqref{eq:event-mechanism} effectively resets $c_i(s)$ to zero, right after an event was triggered.

At the time of this event, the activations of all the units $m \ne \nk$ connected to unit $\nk$ experiences a jump in its state value. 
The jump for $a_{\scx,m}$ is given by:
\begin{align}
    a_{\scx,m}^+(\sk) = a_{\scx,m}^-(\sk) + w_{\scx, m\nk}\,r_{\scx,\nk}\,c_{\nk}^-(\sk)\,,
\end{align}
where $\scx \in \{u,\; r,\; z\}$, $\bfr_\scx = 1$ when $\scx \in \{u, z\}$ and $\bfr_\scx = \bfr$ when $\scx = \{r\}$.
This is equivalent to $y_i = c_{\nk}^-$ being the output of each network unit.
A similar jump is experienced on arrival of an external input, using the appropriate input weights instead (see Supplement for specifics). The event-based asynchronous nature of this update scheme can be formalized in the following proposition.
\begin{proposition}
Let the dynamics of $\bfu(t)$ and $\bfc(t)$ be given by \eqref{eq:current-dynamics} and \eqref{eq:unperturbed-dynamics}. The unperturbed dynamic (in the absence of input) of internal states $\bfc(t)$ are decoupled (the variables of $\bfc(t)$ do not interact with each other).
\label{thrm:decoupled-fwd}
\end{proposition}

\textit{Proof.}
The proof follows directly from the definition \eqref{eq:unperturbed-dynamics}. While variables of single units interact (e.g. $u_i(t)$ and $c_i(t)$), the dynamics of variables of all units are decoupled from each other in the absence of inputs.
\qed

\subsection{Event-based gradient-descent using adjoint method}\label{sec:adjoint-method}

To show that the EGRU gradient updates are event-based, we define the loss over duration $T$ as $\int_{0}^{T} \ell_c(\bfc(t), t)\,dt$, where $\ell_c(\bfc(t), t)$ is the instantaneous loss at time $t$. $T$ is a task-specific time duration within which the training samples are given to the network as events, and the outputs are read out. 
In general $\ell_c(\bfc(t), t)$ may depend arbitrarily on $\bfc(t)$, however in practice we choose the instantaneous loss to depend on the EGRU states only at specific output times to adhere to our fully event-based algorithm.

The loss is augmented with the terms containing the Lagrange multipliers $\bflambda_c,\; \bflambda_{a_\scx}$ to add constraints defining the dynamics of the system from Eqs.~\eqref{eq:current-dynamics},~\eqref{eq:unperturbed-dynamics}. The total loss $\mathcal{L}$ thus reads
\begin{align}\label{eq:full-loss}
    \mathcal{L} = \int_{0}^{T} \left[\ell_c(\bfc(t), t) + \bflambda_c\cdot\left(\tau_{m} \dot{\bfc}(t)-\fun{F}{t, \bfa_u, \bfa_r, \bfa_z,\bfc}\right) + \sum_{\scx\in\{u, r, z\}} \bflambda_{a_\scx}\cdot\left(\tau_{s}\, \dot\bfa_\scx + \bfa_\scx\right)\right] dt \,.
\end{align}
The Lagrange multipliers are referred to as the adjoint variables in this context, and may be chosen freely since both $\tau_{m}\, \dot{\bfc}(t)-\fun{F}{t, \bfa_u, \bfa_r, \bfa_x,\bfc}$ and $\tau_{s}\, \dot\bfa_\scx + \bfa_\scx$ are everywhere zero by construction.

We can choose dynamics and jumps at events for the adjoint variables in such a way that they can be used to calculate the gradient $\frac{d\mathcal{L}}{dw_{ji}}$.
Calculating the partial derivatives taking into account the discontinuous jumps at event times depends on the local application of the implicit function theorem, which requires event times to be a differentiable function of the parameters.
See the Supplement for a full derivation.

The time dynamics of the adjoint variables is given by the following equations with a boundary condition of $\bflambda_{c}(T)=\bflambda_{a_\scx}(T)=0$:
\begin{align}\label{eq:adjoint-dynamics}
   \left(\frac{\partial{F}}{\partial{\bfc}}\right)^T\bflambda_c - \tau_m \dot{\bflambda}_c = 0\,, &\qquad
   \bflambda_{a_\scx}\,+\,\left(\frac{\partial{F}}{\partial{\bfa_\scx}}\right)^T\bflambda_{c} - \tau_s \dot{\bflambda}_{a_\scx} = 0\,,
\end{align}
for $\scx \in \{u, r, z\}$, and $M^T$ denoting the transpose of the matrix $M$. 
The event updates for the adjoints are described in the Supplement.
In practice, the integration of $\bflambda$ is done backwards in time. In analogy to Proposition~\ref{thrm:decoupled-fwd} we find the following property of the forward dynamics.
\begin{proposition}
Let $\bflambda_{c}(t)$ and $\bflambda_{a_\textsc{x}}(t)$ be given by the adjoint dynamics Eq.~\eqref{eq:adjoint-dynamics}. Let the initial conditions of $\bflambda_c$ be given by $\bflambda_c(\sk)$. Then the system of differential equations $\bflambda_c(t)$ for $\skm < t < \sk$ is decoupled (the variables of $\bflambda_c$ do not interact with each other).
\label{thrm:decoupled}
\end{proposition}
\textit{Proof.}
The proof follows from \eqref{eq:adjoint-dynamics}. Since the dynamics of $\bfa_\textsc{x}$ and $\bfc$ are decoupled (unit-wise) between events as shown in Proposition~\ref{thrm:decoupled-fwd}, the dynamics of $\bflambda_c$ are also decoupled. 
The term $\left({\partial{F}}/{\partial{\bfa_\scx}}\right)^T\bflambda_{c}$ reduces to a vector without cross-unit interactions as shown in the Supplement.
\qed

For the recurrent weights $w_{\scx, ij}$ from the different parameter matrices $W_\textsc{x}$ for $\textsc{x} \in {u,r,z}$, we can write the weight updates using only quantities calculated at events $e_k$ as $\Delta w_{\scx, ij} \;=\; \frac{\partial}{\partial w_{\scx, ij}} \mathcal{L}(\bfW)$. We find the following property of the weight updates.
\begin{theorem}
Let $\bflambda_{c}(t)$ and $\bflambda_{a_\textsc{x}}(t)$ be given by the adjoint dynamics Eq.~\eqref{eq:adjoint-dynamics} such that Propositions~\ref{thrm:decoupled-fwd} and \ref{thrm:decoupled} hold. Let $\bfmu_X^-(\sk) = \bfr_\scx^-(\sk) \hp \bfc^{-}(\sk)$. Then the weight updates $\Delta w_{\scx, ij} \;=\; \frac{\partial}{\partial w_{\scx, ij}} \mathcal{L}(\bfW)$ are independent of each other and fully determined by $\bfmu_X^-(\sk)$ and $\bflambda_{a_\textsc{x}}^-(\sk)$, the set of EGRU states evaluated at event times $\sk$, i.e. $\Delta w_{\scx, ij} \;=\; -\tau_s\,\sum_{k=1}^K \bfmu_X^-(\sk) \otimes \bflambda_{a_\textsc{x}}^+(\sk)$.
\label{thrm:independent}
\end{theorem}
\textit{Proof sketch.} Here, $\otimes$ is the outer product, $\bfc^-$ refers to the value of $\bfc(t)$ just before event $e_k$, $\bfr_\textsc{x}^-=0$ for $\textsc{x} \in \{u, z\}$ and equal to the value of $\bfr(t)$ just before event $e_k$ for $\textsc{x} = \{r\}$, $\bflambda_{a_\textsc{x}}^+$ refers to the value of the adjoint variable $\bflambda_{a_\textsc{x}}(t)$ just after the event $e_k$, and $K=|\bfe|$ is the total number of events.
Thus, the values of $\bfr(t),\,\bfc(t)$ need to be stored only at event times, and $\bflambda_{a_\textsc{x}}(t)$ needs to be calculated only at these times, making the gradient updates event-based.
See the Supplement for more details to the proof and for update rules for the input weights and biases. Finally we find the following corollary for the algorithmic complexity of the learning algorithm.

\begin{corollary}
Given $\bfmu_X^-(\sk)$ and $\bflambda_{a_\textsc{x}}^-(\sk)$, the time complexity of parameter update computation $\Delta w_{\scx, ij}$ grows linearly with $K$ (the number of events).
\label{thrm:complexity}
\end{corollary}

\textit{Proof.}
The proof follows from Theorem~\ref{thrm:independent}. The computation required for the outer product $\bfmu_X^-(\sk) \otimes \bflambda_{a_\textsc{x}}^+(\sk)$ depends on the network size and is independent of the total number of events $K$. The sum thus grows linearly with $K$.    
\qed

\todo{Add a summary line here perhaps.}

\section{Results}

\subsection{Gesture prediction}
\label{sec:results-gesture}

\begin{figure}[htbp]
  \centering
    \includegraphics[width=\textwidth]{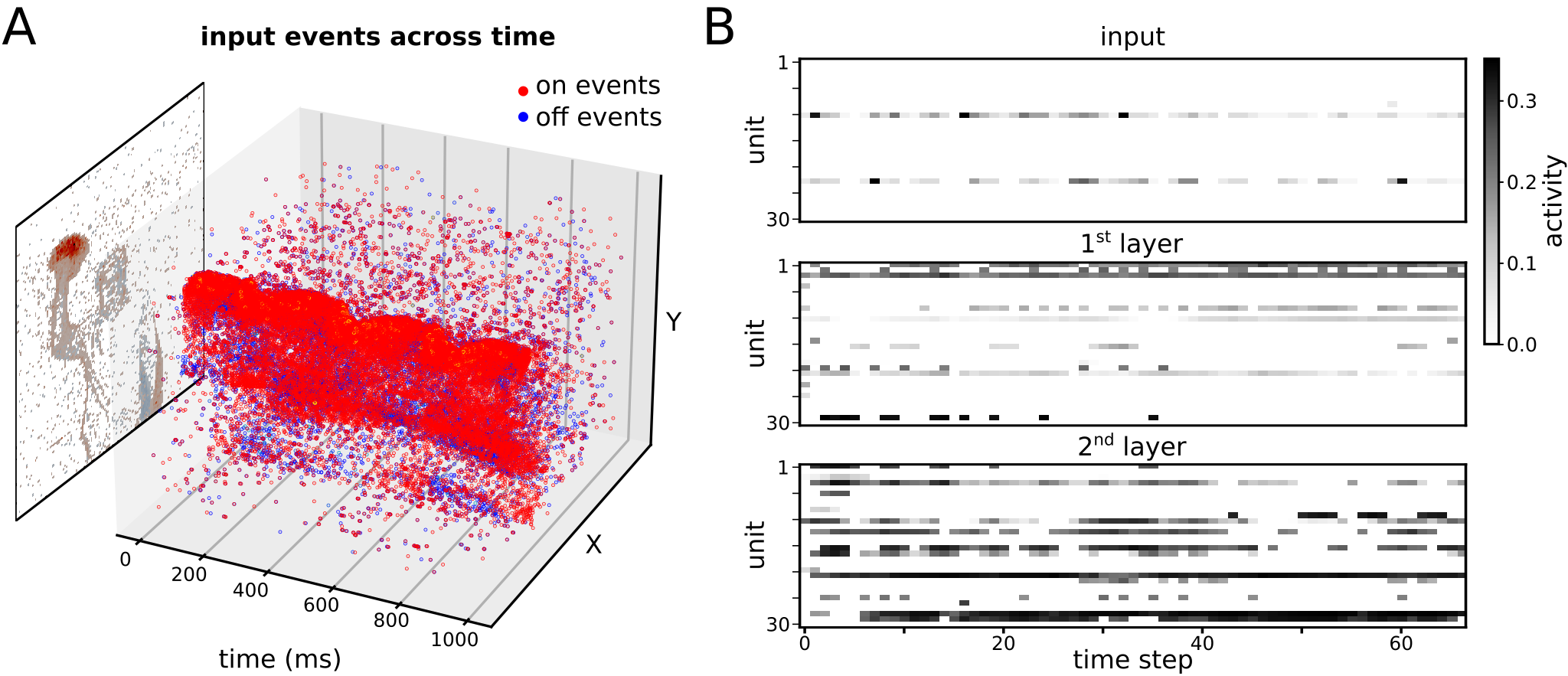}
  \caption{
  \label{fig:gesture}
  \textbf{A:} Illustration of DVS gesture classification data for an example class (right hand wave). On (red) and off (blue) events are shown over time and merged into a summary image for illustration (not presented to the network). \textbf{B:} Sparse activity of input and EGRU units (random subset of 30 units shown for each layer).
  }
  \vspace{-5pt}%
\end{figure}

We evaluated our model on gesture prediction, which is a popular real-world benchmark for RNNs and widely used in neuromorphic research.
The DVS128 Gesture Dataset \citep{amir2017low}, provides sparse event-based inputs which enables us to demonstrate our model's performance and computational efficiency.
The dataset contains 11 gestures from 29 subjects recorded with a DVS128 event camera \citep{lichtsteiner2008128}. 
Each event encodes a relative change of illumination and is given as spatio-temporal coordinates of X/Y position on the $128\times128$-pixel sensor and time stamp. 
We combined the raw event times into `frames' by binning them over time windows of 25\,ms, and then downscaled them to $32\times32$ pixels using a maxpool layer. 
\begin{table}[hbtp]
\centering
\begin{tabular}{@{}c c r r c c c@{}}
  \toprule
          model & hidden & {para-}  & {effective}  & test     & activity & backward  \\
              & dim    & {meters} &  {MAC}       & accuracy & sparsity & sparsity\\
  \midrule
  LSTM~\citep{he2020comparing}  & 512                 & 7.4\si{\million}  & 3.9\si{\million}     & 86.8\% & -        & -        \\
  AlexNet+LSTM+DA & \multirow{2}{*}{256} & \multirow{2}{*}{\tablenum{8.3}\si{\million}} & \multirow{2}{*}{601.3\si{\million}} & \multirow{2}{*}{97.7\%} & \multirow{2}{*}{-}        & \multirow{2}{*}{-}        \\
   \citep{innocenti2021temporal} & & & & & \\
  \midrule
  GRU & 1024           & 15.7\si{\million} & 10.6\si{\million}    & 88.1\% & 0\%      & -        \\

  EGRU & 512           & 5.5\si{\million}  & \bfseries 0.98\si{\million}     & 88.0\% & 83.8\%  & 46.8\%  \\ %
      GRU+DA & 1024           & 15.7\si{\million} & 10.6\si{\million}    & 95.1\% & 0\%      & -        \\

  EGRU+DA & 1024       & 15.7\si{\million} & 3.2\si{\million}    & 97.1\% & 78.8\%  & 58.2\%  \\

      CNN+GRU+DA & 136    & 2.1\si{\million}  &  79.1\si{\million}         & 97.4\% & 0\%  & -      \\

  CNN+EGRU+DA & 795    & 4.8\si{\million}  &  80.6\si{\million}         & \textbf{97.8\%} & 76.4\%  & 72.3\%       \\

  \bottomrule
\end{tabular}
\caption{Model comparison for the DVS Gesture recognition task.  Effective number of MAC operations as described in section \ref{sec:sparse-computation}. %
DA stands for Data Augmentation.
}
\label{tab:dvsgesture}
\vspace{-5pt}%
\end{table}

Comparison of model performance on gesture prediction is presented in Table \ref{tab:dvsgesture}. 
The backward sparsity as described in Section \ref{sec:sparse-computation} was calculated at epoch 100.
EGRU consistently outperformed GRU networks of the same size on this task by a small margin. 
Adding data augmentation (DA) by applying random crop, translation, and rotation, as previously done in \cite{innocenti2021temporal}, increased the performance of pure RNN EGRU architecture to over $97\%$, coming close to state-of-the-art architectures without costly AlexNet pre-processing.
We improved this result with a CNN feature extraction head adapted from ALexNet, DropConnect~\citep{DropConnect} applied to the hidden-to-hidden weights and Zoneout~\citep{zoneout2016}, outperforming \cite{innocenti2021temporal} with far fewer parameters and an order of magnitude reduction in MAC operations.
Further experimental details, ablation studies and statistics over different runs can be found in the supplement sections~\ref{suppl-sec:exp-details-dvs}, \ref{sec:ablation-study} and tables~\ref{tab:suppl-dvsgesture}, \ref{tab:thr-ablation} respectively.

\subsection{Sequential MNIST}
We evaluated the EGRU on the sequential MNIST and permuted sequential MNIST tasks~\citep{le2015simple}, which are widely used benchmarks for RNNs.
In the sequential MNIST task, the MNIST handwritten digits were given as input one pixel at a time, and at the end of the input sequence, the network output was used to classify the digit. 
For the permuted sequential MNIST task, the pixels are permuted using a fixed permutation before being given as input in the same way.
We trained a 1-layer EGRU with 590 units (matching the number of parameters with a 512 unit LSTM).
We did not use any regularisation to increase sparsity in this task, so that we could test how much sparsity, both forward and backward, arises naturally in the EGRU.

In Table~\ref{tab:smnist}, we report the results of discrete-time EGRU along with other state-of-the-art architectures. 
EGRU achieved a task performance comparable to previous architectures while using much fewer operations (more than an 5-fold reduction in effective MAC operations compared to GRU).
Further experimental details, and statistics over different runs can be found in the supplement sections~\ref{suppl-sec:exp-details-smnist} and table~\ref{tab:suppl-smnist} respectively.

\begin{table}[hbtp]
    \centering
    \begin{tabular}{@{}l c c r@{}l r@{}l c c c@{}}
        \toprule
        dataset & model & hidden & \multicolumn{2}{c}{para-}  & \multicolumn{2}{c}{effective}  & test     & activity & backward  \\
            &  & dim    & \multicolumn{2}{c}{meters} &  \multicolumn{2}{c}{MAC}       & accuracy & sparsity & sparsity\\
        \midrule
        \multirow{4}{*}{sMNIST} & coRNN & 256 & 134 &\si{\kilo} & 262&\si{\kilo} & \textbf{99.4\%} & - & - \\
        & \citep{gu2020improving} & 512 & 1&\si{\million} & 1&\si{\million} & 98.9\% & - & - \\
        \cmidrule(lr{1em}){2-8}
        & GRU & 590 & 1&\si{\million} & 1&\si{\million} & 98.8\% & - & - \\
        & EGRU & 590 & 1&\si{\million} & \textbf{226}&\textbf{\si{\kilo}} & 98.3\% & 72.1\% & 27.4\% \\
        \midrule
        \multirow{4}{*}{psMNIST} & coRNN & 256 & 134&\si{\kilo} & {262}&{\si{\kilo}} & \textbf{97.3\%} & - & - \\
        & \citep{gu2020improving} & 512 & 1&\si{\million} & 1&\si{\million} & 95.1\% & - & - \\
        \cmidrule(lr{1em}){2-8}
        & EGRU & 590 & 1&\si{\million} & \textbf{195}&\textbf{\si{\kilo}} & 95.1\% & 82.0\% & 8.4\% \\
        \bottomrule
    \end{tabular}
\caption{
Model comparison on sequential MNIST (sMNIST) and permuted sequential MNIST (psMNIST) task. Top-1 test scores, given as percentage accuracy, where higher is better. coRNN is the model described in \citet{rusch2021coupled}.
}
\label{tab:smnist}
\end{table}

\subsection{Language Modeling}
We evaluated our model on language modeling tasks based on the PennTreebank~\citep{PennTreebank} dataset and the WikiText-2 dataset \citep{WikiText}.
We focused exclusively  on the RNN model in this work, and did not consider techniques such as neural cache models~\citep{NeuralCache}, Mixture-of-Softmaxes~\citep{yang2018breaking} or dynamic evaluation~\citep{DynamicEvaluation}, all of which can be used on top of our model.
A strong baseline for gate-based RNN architectures was established by \cite{merity_Regularizing_2017}.
Similarly, our models consisted of three stacked EGRU cells without skip connections.
DropConnect~\citep{DropConnect} was applied to the hidden-to-hidden weights.
The weights of the final softmax layer were tied to the embedding layer~\citep{WeightTyingInan, WeightTyingPress}.
All experimental details, and statistics over different runs can be found in the Supplement in sections~\ref{suppl-subsec:training-ptb} and Table~\ref{tab:suppl-ptb} respectively.

The results presented in Table~\ref{tab:lm} show that EGRU achieved  performance competitive with AWD-LSTM~\citep{merity_Regularizing_2017}.
At the same time, EGRU inherently exhibited activity sparsity that reduced the required computational operations (calculated analytically).
Another desirable property of EGRU for resource constrained systems is a natural approach to model compression through pruning of least active units.
We discuss our experimental findings for this pruning technique using language modeling as a case study in the Supplement (see Sec.~\ref{suppl-subsec:pruning-ptb}).

\begin{table}[htbp]
    \centering
    \sisetup{detect-weight=true,detect-inline-weight=math}
    \begin{tabular}{@{}l c c c S[table-format=2.1] c c c@{}}
      \toprule
      dataset & model    & hidden   & para-  & {effective}        & test & activity & backward \\
         &      & dim         & meters & {MAC}              &  & sparsity & sparsity \\
      \midrule
      \multirow{5}{*}{PTB} & \citet{merity_Regularizing_2017} & 1150 & 24\si{\million} & 24.0\si{\million}  & 57.3 & -        & -        \\
      & \citet{li_Independently_2018}    & -               & 52\si{\million} & 51.6\si{\million}               & \textbf{55.2} & -        & -        \\
      \cmidrule(lr{1em}){2-8}
      & GRU          & 1350         & 28\si{\million}   & 27.6\si{\million}  & 66.3 & -        & -        \\
      & EGRU         & 1350         & 33\si{\million}   & \bfseries 9.9\si{\million}   & 57.2 & 79.7\%  & 46.3\%  \\
      & EGRU         & 2000         & 55\si{\million}   & 12.7\si{\million}  & 57.0 & 84.8\%  & 42.9\%  \\
      \midrule
      \multirow{6}{*}{WikiText-2} & \cite{Melis2018}                & -      & 24\si{\million}    & {-}         & 65.9 & -        & -        \\
      & \cite{merity_Regularizing_2017} & 1150   & 33\si{\million}    & 32.0\si{\million}       & \textbf{65.8} & -        & -        \\
      \cmidrule(lr{1em}){2-8}
      & GRU                      & 1350 & 39\si{\million} & 38.8\si{\million} & 71.8 & - & - \\
      & {EGRU}                   & 1350 & 51\si{\million} & \bfseries 17.5\si{\million} & 70.6 & 76.8~\% & 44.2~\% \\
      & {EGRU}                   & 2000 & 74\si{\million} & 20.1\si{\million} & 68.9 & 82.7~\% & 42.2~\% \\
      \bottomrule
    \end{tabular}
    \caption{Model comparison on PennTreebank and WikiText-2. Test scores are given as perplexities, where lower is better. 
    Effective MAC operations are given for a single time step and consider the layer-wise sparsity in the forward pass. 
    Activity sparsity is given for the trained model to resemble inference sparsity. 
    Backward sparsity is averaged over the whole training.
    Model parameters were optimized on Penn Treebank and transfered to WikiText-2.}
    \label{tab:lm}
\vspace{-5pt}%
\end{table}

\section{Discussion}
\label{sec:discussion}
This work introduces the EGRU, a new form of a recurrent neural network that uses a biologically inspired activity sparsity mechanism.
The EGRU extends the GRU with an event-generating mechanism using a threshold and uses surrogate gradients to train the model with BPTT, which leads to computationally sparse forward and backward passes.
We theoretically proved that in the continuous time limit, the model's inference and training scales proportionally to the number of events generated and the number of neurons, but not with the number parameters as in conventional RNN models.

The EGRU achieved competitive task performance on tasks such as gesture recognition, sequential image classification and language modeling while achieving activity sparsity of up to 85\% ($15\%$ of the units active on average). 
Scaling up networks for language modeling has shown some of the most promising results in the last few years~\citep{brown_Language_2020}, hence, our choice of this task, albeit on a smaller scale.
Considering the need for extensive hyperparameter search \citep{Melis2018} for language modeling, our model achieved promising results while maintaining a high degree of activity-sparsity.
For example, our EGRU with 1350 hidden units reached perplexities better than reported by LSTM and GRU baselines, while maintaining the high level of activity-sparsity.
To the best of our knowledge, this is the first demonstration of such activity sparsity mechanisms that yields strong benchmark performance compared to baselines.

While we base our model on the GRU due to its simplicity, this formulation can easily be extended to any arbitrary network dynamics, including the LSTM, allowing specialized architectures for different domains.
The adjoint method for hybrid systems that we use for analysis here can also be used as a powerful general-purpose tool to formulate cotinuous time models that are activity-sparse and trained using event-based gradient descent updates for any recurrent neural network architectures.
Another novel outcome of this paper is that this theory can handle inputs in continuous time as events, which is very intuitive, hence providing an alternative to the more complex controlled differential equations~\citep{kidger_Neural_2020a}.
The EGRU can also be used for irregularly spaced sequential data quite naturally.

The compute efficiency of this model can directly translate into gains in energy efficiency when implemented using event-based software primitives.
The model will work well on heterogeneous compute resources, including pure CPU nodes, neuromorphic devices such as Intel's Loihi~\citep{davies2018loihi} and SpiNNaker 2~\citep{hoppner2017dynamic}, that can achieve orders of magnitude higher energy efficiency, as well as on deep learning hardware that support dynamic sparsity, such as the Graphcore system \citep{jia2019dissecting}.
On neuromorphic devices with on-chip memory in the form of a crossbar array, the activity sparsity directly translates into energy efficiency. 
For larger models that need off-chip memory, activity-sparsity needs to be combined with parameter-sparsity to reduce energy-intensive memory access operations.

In summary, with the motivation of building scalable, energy-efficient deep recurrent models, we presented an activity sparsity mechanism that reduces the required compute for both inference and learning. 
We demonstrated that the EGRU -- our GRU-based model using this activity sparsity mechanism -- is a competitive alternative to GRU and LSTM models, especially for resource constrained systems and neuromorphic devices. 
In future work, we plan to combine this activity sparsity mechanism with methods for parameter sparsity and learning long-range dependencies  and implement them on neuromorphic hardware to realise the efficiency gains and scalability of recurrent neural network architectures.

\section{Reproducibility}
We ensure that the results presented in this paper are easily reproducible using just the information provided in the main text as well as the appendix. Details of the models used in our experiments are presented in the main paper and further elaborated in the appendix. We provide additional experimental details, ablation studies, and statistics over multiple runs in the appendix Section \ref{suppl-sec:exp-details}. We use publicly available libraries (Appendix section~\ref{suppl-sec:hardware-software}) and Datasets (Appendix section~\ref{suppl-sec:dataset-licenses}) in our experiments. We will further provide the source code to the reviewers and ACs in an anonymous repository once the discussion forums are opened. The included code will also contains ''readme'' texts to facilitate easy reproducibility. The theoretical analysis provided in Section \ref{sec:theoritical-analysis} is derived in the appendix along with the event-based learning rule.

\section{Acknowledgements}
The authors gratefully acknowledge the GWK support for funding this project by providing computing time through the Center for Information Services and HPC (ZIH) at TU Dresden.
We acknowledge the use of Fenix Infrastructure resources, which are partially funded from the European Union's Horizon 2020 research and innovation programme through the ICEI project under the grant agreement No. 800858. 
AS was funded by the Ministry of Culture and Science of the State of North Rhine-Westphalia, Germany during this work.
KKN is funded by the German Federal Ministry of Education and Research (BMBF) within the KI-ASIC project (16ES0996).
MS is fully funded by a grant of the Bosch Research Foundation.
CM receives funding from the German Research Foundation (DFG, Deutsche Forschungsgemeinschaft) as part of Germany’s Excellence Strategy – EXC 2050/1 – Project ID 390696704 – Cluster of Excellence “Centre for Tactile Internet with Human-in-the-Loop” (CeTI) of Technische Universität Dresden.
DK is funded by the German Federal Ministry of Education and Research (BMBF) within the project EVENTS (16ME0733).
The authors would like to thank Darjan Salaj, Melika Payvand, Markus Murschitz, Franz Scherr, Robin Schiewer for helpful comments on this manuscript.
AS would also like to thank Darjan Salaj and Franz Scherr for insightful early discussions, and Laurenz Wiskott for institutional support.

\bibliography{references}

\makeatletter\@input{yy.tex}\makeatother

\end{document}

\maketitle
\setcounter{table}{0}
\renewcommand{\thetable}{S\arabic{table}}%
\setcounter{figure}{0}
\renewcommand{\thefigure}{S\arabic{figure}}%
\setcounter{equation}{0}
\renewcommand{\theequation}{S\arabic{equation}}

\appendix

\section{Derivation of continuous time version of GRU}

In this section we derive the continuous-time version of the GRU model. 
Note that our definition of the GRU differs from the original version, presented in \citep{cho2014learning}, by inverting the role of the $\bfu\dti$ and $1-\bfu\dti$ terms in the updates equations for $\bfy\dti$ (a change in sign). This substitution does not change the behavior of the model but simplifies the notation in the continuous-time version of the model.

We first rewrite the dynamics of a layer of GRU units at time step $t$ from Eq.~\eqref{eq:gru-dt} of the main text, separating out the input and recurrent weights: 
\begin{equation}
\label{s-eq:gru-dt}
\begin{split}
\bfu\dti = \fun{\sigma}{ \bfU_{u}  \bfx\dti + \bfV_{u} \bfy\dtim1 + \bfb_u }\,, & \quad
     \bfr\dti = \fun{\sigma}{\bfU_{r} \bfx\dti + \bfV_{r} \bfy\dtim1 + \bfb_r }\,, \\[2mm]
     \bfz\dti = \fun{g}{ \bfU_{z} \bfx\dti +  \bfV_{z} \left(\bfr\dti \hp\; \bfy\dtim1\right) + \bfb_z }\,, & \quad
     \bfy\dti = \bfu\dti \hp\; \bfz\dti  + (1-\bfu\dti) \hp\; \bfy\dtim1\,,
\end{split}
\end{equation}
we can write this as 
\begin{equation}
    \label{s-eq:gru-dt-re}
    \bfy\dti - \bfy\dtim1 = - \bfu\dti \hp \bfy\dtim1 + \bfu\dti \hp \bfz\dti.
\end{equation}
Note that $\bfu$ here is equivalent to $\tilde\bfu = 1 - \bfu$ used in the standard GRU model.
Eq.~\eqref{s-eq:gru-dt-re} is in the form of a forward Euler discretization of a continuous time dynamical system.
Defining $\bfy(t) \equiv \bfy\dtim1$, we get $\bfr(t) \equiv \bfr\dti, \bfu(t) \equiv \bfu\dti, \bfz(t) \equiv \bfz\dti$.
Let $\Delta t$ define an arbitrary time step. Then Eq.~\eqref{s-eq:gru-dt-re} becomes:
\begin{equation}
    \bfy(t+\Delta t) - \bfy(t) = -\bfu(t) \hp \left(\bfy(t) + \bfz(t)\right) \Delta t
\end{equation}
Dividing by $\Delta t$ and taking limit $\Delta t \rightarrow 0$, we get:
\begin{equation}
    \dot{\bfy}(t) = -\bfu(t) \hp \left(\bfy(t) - \bfz(t)\right),
\end{equation}
where $\dot\bfy(t) \equiv \frac{d\bfy(t)}{dt}$ is the time derivative of $\bfy(t)$.

\section{Full details of the continuous time EGRU}

In this section we establish the continuous time version of the EGRU model. To describe the event generating mechanism and state dynamics it is convenient to express the dynamical system equations in therms of the activations $\bfa_\scx$.

We first rewrite Eqs.~\eqref{eq:current-dynamics} \&~\eqref{eq:unperturbed-dynamics} of the main text, as:
\begin{align}
    f_{a_\scx} &\quad\equiv\quad \tau_{s}\, \dot\bfa_\scx + \bfa_\scx\ + \bfb_\scx \,=\, \mathbf{0}\,, \quad \scx \in \{u,\; r,\; z\} \\
    f_c &\quad\equiv\quad \tau_{m}\, \dot{\bfc}(t)  +  \bfu(t) \hp \left(\bfc(t) - \bfz(t) \right) \;=\; \tau_{m}\, \dot{\bfc}(t) - \fun{F}{t, \bfa_u, \bfa_r, \bfa_z,\bfc} = 0\,.
\end{align}

We write the event transitions for $\bfc$ at network event $\ek \in \bfe$, $\ek = \bwrap{\sk,\nk}$, where $\sk$ are the continuous (real-valued) event times, and $\nk$ denotes which unit got activated, 
and using the superscript $.^-$ ($.^+$) to the quantity just before (after) the event, as:
\begin{equation}\label{s-eq:c-jumps}
    c_{\nk}^-(\sk) = \vartheta_{\nk}\,, \qquad c_{\nk}^+(\sk) = 0\,, \qquad c_m^+(\sk)=c_m^-(\sk).
\end{equation}
where $m \ne \nk$ denotes all the units connected to unit $\nk$ that are not activated.
At the time of this event, the activations $a_{\scx,m}$ ($\scx \in \{u, r, x\}$) experiences a jump in its state value, given by:
\begin{align}
    \label{s-eq:amu-jumps}
    a_{u,m}^+(\sk) &= a_{u,m}^-(\sk) + v_{u, m\nk}\,\times\,c_{\nk}^-(\sk)\,, \\
    \label{s-eq:amr-jumps}
    a_{r,m}^+(\sk) &= a_{r,m}^-(\sk) + v_{r, m\nk}\,\times\,c_{\nk}^-(\sk)\,,\\
    \label{s-eq:amz-jumps}
    a_{z,m}^+(\sk) &= a_{z,m}^-(\sk) + v_{z, m\nk}\,\times\,r_{\nk}\,\times\,c_{\nk}^-(\sk)\,,\\
    \label{s-eq:an-jumps}
    a_{\scx,\nk}^+(\sk) &= a_{\scx,\nk}^-(\sk)\,.
\end{align}

External inputs also come in as events $\tek \in \tilde{\bfe}$, $\tilde{\ek} = \bwrap{\sk,\ik}$, where $\sk$ are the continuous (real-valued) event times, and $\ik$ denotes the index of the input component that got activated. 
Only the activations $a_{\scx,l}$ for the $l$-th unit experience a transition/jump on incoming external input events, as follows:
\begin{align}
    a_{\scx,l}^+(\sk) &= a_{\scx,l}^-(\sk) + u_{\scx, l\nk}\,\times\,x_{\ik}(\sk)\,,
\end{align}
where $x_{\ik}(\sk)=\left(\bfx(\sk)\right)_{\ik}$ is the $\ik$-th component of the input $\bfx$ at time $\sk$.
The internal state $\bfc$ remains the same on the external input event. That is, $c_l^+=c_l^-$.

\section{Details for Proof of \textbf{Proposition 2}}

Using basic matrix algebra, it can be shown that both $\frac{\partial{F}}{\partial{\bfc}}$ and $\frac{\partial{F}}{\partial{\bfa_\scx}}$ simplify to a diagonal matrix due to the independence of a notional unit $i$ from unit $j$ in the forward dynamics in Eqns.~\eqref{eq:current-dynamics}, \eqref{eq:unperturbed-dynamics}.
Therefore, Eqn.~\eqref{eq:adjoint-dynamics} can be written as the following for unit $i$:
\begin{align}
    \frac{\partial \dot{c}_i}{\partial c_i} \lambda_{c,i} - \tau_m \dot{\lambda}_{c,i} = 0\,, &\qquad 
    \lambda_{a_\scx,i} + \frac{\partial \dot{c}_i}{\partial a_{\scx,i}} \lambda_{c,i} - \tau_s \dot{\lambda}_{a_\scx, i} = 0\,,
\end{align}
where $\dot{c}_i = (F)_i$, the $i$th element of $F$.

\section{Derivation of event-based learning rule in continuous time}

In this section we derive the event-based updates for the network weights. The update questions yield different results for the recurrent weights ($\bfV_\scx$), biases ($\bfb_\scx$) and input weights ($\bfU_\scx$), which are derived in the remainder of this section. To increase readability important terms are highlighted in color.

\subsection{Gradient updates for the recurrent weights $\bfV_\scx$: Proof of \textbf{Theorem 1}}
We first split the integral Eq.~\eqref{eq:full-loss} across events as:
\begin{align}
    \cL = \sum_{k=0}^N\int_{\sk}^{\skp} \left[\ell_c(\bfc(t), t) + \bflambda_c\,\cdot\,f_c + \sum_{\scx\in\{u, r, z\}} \bflambda_{a_\scx}\,\cdot\,f_{a_\scx}\right] \, dt \;.
\end{align}
Then taking the derivative of the full loss function, we get:
\begin{align}\label{s-eq:full-integral}
    \dbyd{\cL}{v_{ji}} = \dbyd{}{v_{ji}} \left\{  \sum_{k=0}^N\int_{\sk}^{\skp} \left[\textcolor{Apricot}{\ell_c(\bfc(t), t)} + \textcolor{Aquamarine}{\bflambda_c\,\cdot\,f_c} + \textcolor{Bittersweet}{\sum_{\scx\in\{u, r, z\}} \bflambda_{a_\scx}\,\cdot\,f_{a_\scx}}\right] \, dt  \right\}\;.
\end{align}

By application of Leibniz integral rule we get,
\begin{align}
    \color{Apricot}
    \dbyd{}{v_{ji}} \int_{\sk}^{\skp} \ell_c(\bfc(t), t) dt &= \ell_c(\bfc, \skp) \dbyd{\skp}{v_{ji}} - \ell_c(\bfc, \sk) \dbyd{\sk}{v_{ji}} + \int_{\sk}^{\skp} \dobydo{\ell_c}{\bfc} \cdot \dobydo{\bfc}{v_{ji}} dt \;.
\end{align}
and
\begin{align}
     &\textcolor{Aquamarine}{\dbyd{}{v_{ji}} \int_{\sk}^{\skp}\bflambda_c\,\cdot\,f_c \,dt}  \\ &\;=\; \int_{\sk}^{\skp}\bflambda_c\,\cdot\,\dbyd{f_c}{v_{ji}} \,dt \;=\; \int_{\sk}^{\skp}\bflambda_c\,\cdot\,\left\{\tau_m \dbyd{}{t}\dobydo{\bfc}{v_{ji}} + \dobydo{F}{v_{ji}}\right\} \,dt \\
    &\;=\; \tau_m\, \left[\bflambda_c\,\cdot\,\dobydo{\bfc}{v_{ji}}\right]_{\sk}^{\skp} \\
                                                                  &\qquad -\; \tau_m \int_{\sk}^{\skp} \left\{
                                                                      \dot{\bflambda}_c\,\cdot\,\dobydo{\bfc}{v_{ji}} 
                                                                      + \bflambda_c \,\cdot\, \left(\left(\dobydo{F}{\bfc}\right)^T\dobydo{\bfc}{v_{ji}} 
                                                              +  \sum_{\scx \in \{u, r, z\}}  \left(\dobydo{F}{\bfa_\scx}\right)^T \dobydo{\bfa_\scx }{v_{ji}}\right)\right\} \,dt \;,
\end{align}
where we first apply Gronwall's theorem~\cite{gronwall1919note}, then integration by parts, and $M^T$ denotes the transpose of matrix $M$.
$\ell_c(\bfc(t), t)$ is the instantaneous loss evaluated at time $t$.
Similarly,
\begin{align}
    \color{Bittersweet}
    \dbyd{}{v_{ji}} \int_{\sk}^{\skp}\sum_{\scx\in\{u, r, z\}} \bflambda_{a_\scx}\,\cdot\,f_{a_\scx} \,dt &= \sum_{\scx\in\{u, r, z\}} \int_{\sk}^{\skp} \bflambda_{a_\scx}  \,\cdot\, \left\{ \tau_s \dbyd{}{t} \dobydo{\bfa_\scx}{v_{ji}} + \dobydo{\bfa_\scx}{v_{ji}}\right\} \,dt \\
                                                                                                          &= \tau_s\, \left[\bflambda_{a_\scx} \,\cdot\, \dobydo{\bfa_\scx}{v_{ji}}\right]_{\sk}^{\skp} - \tau_s  \int_{\sk}^{\skp} \left\{ \dot{\bflambda}_{a_\scx} \,\cdot\, \dobydo{\bfa_\scx}{v_{ji}} + \bflambda_{a_\scx} \,\cdot\, \dobydo{\bfa_\scx}{v_{ji}}\right\} \,dt \;,
\end{align}
since $\dobydo{\bfb}{v_{ji}} = 0$.

Substituting these values into Eq.~\eqref{s-eq:full-integral}, and setting the coefficients of terms with $\dobydo{\bfc}{v_{ji}} \text{and} \dobydo{\bfa_\scx}{v_{ji}}$ to zero (using the fact that we can choose the adjoint variables freely due to $f_c$ and $f_{a_\scx}$ being everywhere zero by definition), we get the dynamics of the adjoint variable described in Eq.~\eqref{eq:adjoint-dynamics}.
The adjoint variable is usually integrated backwards in time starting from $t=T$, also due to its dependence on the loss values.
The initial conditions for the adjoint variables is defined as $\bflambda_c = \bflambda_{a_\scx} = \mathbf{0}$.

Setting the coefficients of terms with $\dobydo{\bfc}{v_{ji}} \text{and} \dobydo{\bfa_\scx}{v_{ji}}$ to zero allows us to write the parameter updates as:
\begin{align}
    \label{s-eq:xi}
    \dbyd{\cL}{v_{ji}} &= \sum_{k=0}^N \left\{ \left( l_c^- - l_c^+ \right) \dbyd{s}{\vji} + \tau_s \sum_\scx \left( \bflambda_{a_\scx}^- \cdot \dobydo{\bfa_\scx^-}{\vji} - \bflambda_{a_\scx}^+ \cdot \dobydo{\bfa_\scx^+}{\vji}\right) + \tau_m \left( \bflambda_c^- \cdot \dobydo{\bfc^-}{\vji} - \bflambda_c^+ \cdot \dobydo{\bfc^+}{\vji}\right) \right\} \\
                       &= \sum_{k=0}^N \xi_{\scx, ijk}
\end{align}

To define the required jumps at event times for the adjoint variables, we start with finding the relationship between $\dobydo{\bfc^-}{v_{ji}} \text{and} \dobydo{\bfc^+}{v_{ji}}$.
Eqs.~\eqref{s-eq:c-jumps} define $\sk$ as a differentiable function of $\vji$ under the condition $\dot{c}_{\nk}^- \neq 0$ and $\dot{c}_{\nk}^+ \neq 0$ due to the implicit function theorem~\citep{wunderlich_Eventbased_2021,yang2014proof}.
\begin{align}
    c_{\nk}^- - \vartheta_\nk &= 0 \\
    \dobydo{c_{\nk}^-}{\vji} + \dbyd{c_{\nk}^-}{s} \dobydo{s}{\vji} &= 0 \\
    \dobydo{c_{\nk}^-}{\vji} + \dot{c}_{\nk}^- \dobydo{s}{\vji} &= 0 \\
    \dobydo{s}{\vji} &= \frac{-1}{\dot{c}_{\nk}^-} \dobydo{c_{\nk}^-}{\vji} \;,
\end{align}
where we write $\dbyd{c_{\nk}^-}{s} \equiv \dot{c}_{\nk}^-$ and $\dot{c}_{\nk}^- \neq 0$.
Similarly, 
\begin{align}
    c_{\nk}^+ &= 0\\
    \dobydo{c_{\nk}^+}{\vji} + \dot{c}_{\nk}^+ \dobydo{s}{\vji} &= 0
\end{align}
which allows us to write 
\begin{align}
    \label{s-eq:docn}
    \dobydo{c_{\nk}^+}{\vji} = \frac{\dot{c}_{\nk}^+}{\dot{c}_{\nk}^-} \dobydo{c_{\nk}^-}{\vji}
\end{align}

Similarly, starting from $c_m^+ = c_m^-$, we can derive
\begin{align}
    \label{s-eq:docm}
    \dobydo{c_m^+}{\vji} = \dobydo{c_m^-}{\vji} - \frac{1}{\dot{c}_{\nk}^-} \dobydo{c_{\nk}^-}{\vji} \left(\dot{c}_m^- - \dot{c}_m^+\right)
\end{align}

For the activations $\bfa_\scx$, we use Eqs.~\eqref{s-eq:amu-jumps}--\eqref{s-eq:an-jumps} to derive the relationships between $\dobydo{a_{\scx}}{\vji}^+$ and $\dobydo{a_{\scx}}{\vji}^-$.
Thus, we have:
\begin{align}
    \label{s-eq:doam}
    \dobydo{a_{\scx,m}^+}{\vji} &= \dobydo{a_{\scx,m}^-}{\vji} - \frac{1}{\tau_s} \frac{v_{m\nk}\,r_{\scx,\nk}^- c_{\nk}^-}{\dot{c}_{\nk}^-} \dobydo{c_{\nk}^-}{\vji} + \delta_{i\nk} \delta_{jm} c_{\nk}^- + c_\nk^- v_{m\nk} \dobydo{r_{\scx,\nk}^-}{\vji} - c_\nk^- v_{m\nk} \frac{\dot{r}_{\scx,\nk}^-}{\dot{c}_\nk^-} \dobydo{c_\nk^-}{\vji} \\
    \label{s-eq:doan}
    \dobydo{a_{\scx,\nk}^+}{\vji} &= \dobydo{a_{\scx,\nk}^-}{\vji}
\end{align}
where $\bfr_\scx = \mathbf{0}$ if $\scx \in \{u, r\}$ and $\bfr_\scx = \bfr$ if $\scx = \{ z \}$.

Substituting Eqs.~\eqref{s-eq:docn}, \eqref{s-eq:docm}, \eqref{s-eq:doan}, \eqref{s-eq:doam} into Eq.~\eqref{s-eq:xi}, we get:
\begin{align}
    \xi_{\scx,ijk} = \left\{ 
        \dobydo{c_\nk^-}{\vji} \left(
            \frac{-1}{\dot{c}_\nk^-} \left(\ell_c^+ - \ell_c^- \right) + \tau_m \left(\lambda_{c,\nk}^- - \frac{\dot{c}_\nk^+}{\dot{c}_\nk^-} \lambda_{c, \nk}^+\right)
                + \tau_m \frac{1}{\dotcnk^-} \sum_{m \neq \nk} \lambda_{c,m}^+ \left(\dotcm^- - \dotcm^+ \right) \right. \right. \\
                \left. \left. + \sum_\scx \frac{r_{\scx,\nk}^- c_\nk^-}{\dotcnk^-} \sum_{m \neq \nk} v_{m\nk} \lambda_{a_\scx, m}^+
                + \tau_s \sum_\scx \frac{\dot{r}_{\scx,\nk}^- c_\nk^-}{\dotcnk^-} \sum_{m \neq \nk} v_{m\nk} \lambda_{a_\scx, m}^+
            \right) \right. \\
            \left. + \tau_m \sum_{m \neq \nk} \dobydo{c_m^-}{\vji} \left( \lambda_{c,m}^- - \lambda_{c,m}^+ \right) \right. \\
            \left. \tau_s \sum_\scx \dobydo{a_{\scx, \nk}^-}{\vji} \left( \left( \lambda_{a_\scx,\nk}^- - \lambda_{a_\scx,\nk}^+ \right) - c_\nk^- G'(a_{\scx, \nk}^-) \sum_{m \neq \nk} v_{m\nk} \lambda{a_\scx, m}^+  \right) \right. \\
            \left. \tau_s \sum_\scx \sum_{m \neq \nk} \dobydo{a_{\scx,m}^-}{\vji} \left(\lambda_{a_\scx,m}^- - \lambda_{a_\scx,m}^+ \right)\right. \\
            \left. -\tau_s \delta_{i\nk} r_{\scx, \nk}^- c_\nk^- \sum_{m \neq \nk} \delta_{jm} \lambda_{a_\scx,m}^+\right\}
\end{align}
where we use $\bfr_\scx = G(\bfa_\scx)$ to denote $G(\bfa_r) = \bfr$ and $G(\bfa_z)=G(\bfa_u)=\mathbf{1}$, $\delta_{ab}$ is the kronecker delta defined as:
\begin{equation}
    \delta_{ab} = \left\{ \begin{array}{rcl}
            1 & \mbox{if} & a = b,\\
            0 & \mbox{otherwise} &
        \end{array}\right.
\end{equation}

Setting the coefficients of $\dobydo{c^-}{\vji}$ and $\dobydo{a_\scx^-}{\vji}$ to $0$ (again, using our ability to choose the adjoint variables freely), we can get both $\xi_{\scx, ijk}$ and the transitions for the adjoint variables.

For the parameter updates we get:
\begin{align}
    \xi_{ijk} &= - \tau_s \delta_{i\nk} r_{\scx,\nk}^- c_\nk^- \sum_{m \neq \nk} \delta_{jm} \lambda_{a_\scx, m}^+ \\
    &= -\tau_s r_{\scx,i}^- c_i^- \lambda_{a_\scx, j}^+ \;.
\end{align}

Thus we can write:
\begin{align}
    \Delta w_{\scx, ij} \;=\; \frac{\partial}{\partial w_{\scx, ij}} \mathcal{L}(\bfW) \;=\; \sum_k \xi_{\scx, ijk}\,.
\end{align}
The corresponding value of $\xi_{\scx, ijk}=(\bfxi_{\scx, k})_{ij}$ is given by the following formula, written in vector form for succinctness:
\begin{align}
    \bfxi_{\scx, k} \;=\; -\tau_s\,\bwrap{\bfr_\scx^-(\sk) \hp \bfc^{-}(\sk)} \otimes \bflambda_{a_\scx}^+(\sk)\;,
\end{align}

The jumps/transitions of the adjoint variables are:
\begin{align}
    \lambda_{a_\scx,m}^+ &= \lambda_{a_\scx,m}^- \\
    \lambda_{a_\scx,\nk}^+ &= \lambda_{a_\scx,\nk}^- - c_\nk^- G'(a_{\scx,\nk}) \sum_{m \neq \nk} v_{m\nk} \lambda_{a_\scx, m}^+ \\
    \lambda_{c,m}^+ &= \lambda_{c,m}^- \\
    \tau_m \dotcnk^+ \lambda_{c,\nk}^+ &= -(\ell_c^+ - \ell_c^-) + \tau_m \dotcnk^- \lambda_{c,\nk}^- + \tau_m \sum_{m \neq \nk} \lambda_{c, m}^+ (\dotcm^- - \dotcm^+) \nonumber \\
    & \qquad + \tau_s c_\nk^- \sum_\scx \left( \dot{r}_{\scx, \nk}^- +  \frac{r_{\scx, \nk}^-}{\tau_s} \right) \sum_{m \neq \nk} v_{m\nk} \lambda_{a_\scx, m}^+ \;,
 \end{align}
where $(\ell_c^+ - \ell_c^-)$ denotes the jumps in the instantaneous loss around event time $s_k$.
Thus, all the quantities on the right hand side of Eq.~\eqref{s-eq:xi} can be calculated from known quantities.

\subsection{Gradient updates for biases $\bfb_\scx$}
Proceeding similarly for the biases $\bfb_\scx$ for each of $\scx \in \{u, r, z\}$ (dropping the subscript $\scx$ for simplicity):
\begin{align}
    \dbyd{\cL}{b_{i}} = \dbyd{}{b_{i}} \left\{  \sum_{k=0}^N\int_{\sk}^{\skp} \left[{\ell_c(\bfc(t), t)} + {\bflambda_c\,\cdot\,f_c} + {\sum_{\scx\in\{u, r, z\}} \bflambda_{a_\scx}\,\cdot\,f_{a_\scx}}\right] \, dt  \right\}\;.
\end{align}
the $\xi_{\scx,ik}^\text{bias}$ term can be shown to be:
\begin{equation}
    \xi_{\scx,ik}^\text{bias} = \int_\sk^\skp \lambda_{a_\scx,i} \,dt
\end{equation}
with
\begin{equation}
    \dbyd{\cL}{b_i} = \sum_{k=0}^N \xi_{\scx,ik}^\text{bias}\;.
\end{equation}

\subsection{Gradient updates for input weights $\bfU_\scx$}
Proceeding similarly for the input weights $\bfU_\scx$ for each of $\scx \in \{u, r, z\}$ (dropping the subscript $\scx$ for simplicity):
\begin{align}
    \dbyd{\cL}{u_{jx}} = \dbyd{}{u_{jx}} \left\{  \sum_{k=0}^N\int_{\sk}^{\skp} \left[{\ell_c(\bfc(t), t)} + {\bflambda_c\,\cdot\,f_c} + {\sum_{\scx\in\{u, r, z\}} \bflambda_{a_\scx}\,\cdot\,f_{a_\scx}}\right] \, dt  \right\}\;.
\end{align}
the $\xi_{\scx,jxk}^\text{input}$ term can be shown to be:
\begin{equation}
    \xi_{\scx,jxk}^\text{input} = - \tau_s \lambda_{a_\scx,j}^+ x_x
\end{equation}
with
\begin{equation}
    \dbyd{\cL}{u_{jx}} = \sum_{k=0}^N \xi_{\scx,jxk}^\text{input}\;.
\end{equation}

\section{Details of experiments}
\label{suppl-sec:exp-details}
\todo[inline]{Talk about EGRUD threshold initialisation, other parameter initialisation}

\subsection{DVS128 Gesture recognition}
\label{suppl-sec:exp-details-dvs}
In this experiment we use Tonic library \citep{tonic_pytorch} to prepare the dataset.
The recordings in the dataset are sliced by time without any overlap to produce samples of length 1.7 seconds.
The data is denoised with a filter time of 10ms and normalised to [0;1] before being fed to the model.
The positive and negative polarity events are represented by 2 separate channels. 
Our model consists of a preprocessing layer which performs downscaling and flattening transformations, followed by two RNN layers. Both RNN layers have the same number of hidden dimensions. 
Finally, a fully connected layer of size 11 performs the classification.
All the weights were initialised using Xavier uniform distribution, while the biases were initialised using a uniform distribution.
The unit thresholds were initialised using a normal distribution with mean 0 and standard deviation of $\sqrt{2}$, but was transformed to their absolute value after every update.
We use cross-entropy loss and Adam optimizer with default parameters (0.001 learning rate, $\beta_1 = 0.9$, $\beta_2=0.999$).
The learning rate is scaled by 80\% every 100 epochs.

We use additional loss to regularize the output and increase sparsity of the network. The applied regularization losses are shown in Eq.~\eqref{eq:dvs-loss}.
$L_{reg}$ is applied indirectly to the active outputs and $L_{act}$ is applied on the auxiliary internal state $c_i\dti$, the threshold parameter $\vartheta_i$ is detached from the graph in the second equation so the loss only affects the internal state.
We set the regularization weights $w_{reg}$ and $w_{v}$ to 0.01 and 0.05 respectively.

\begin{figure}
  \begin{subfigure}{0.5\linewidth}
    \centering
    \includegraphics[width=0.95\linewidth]{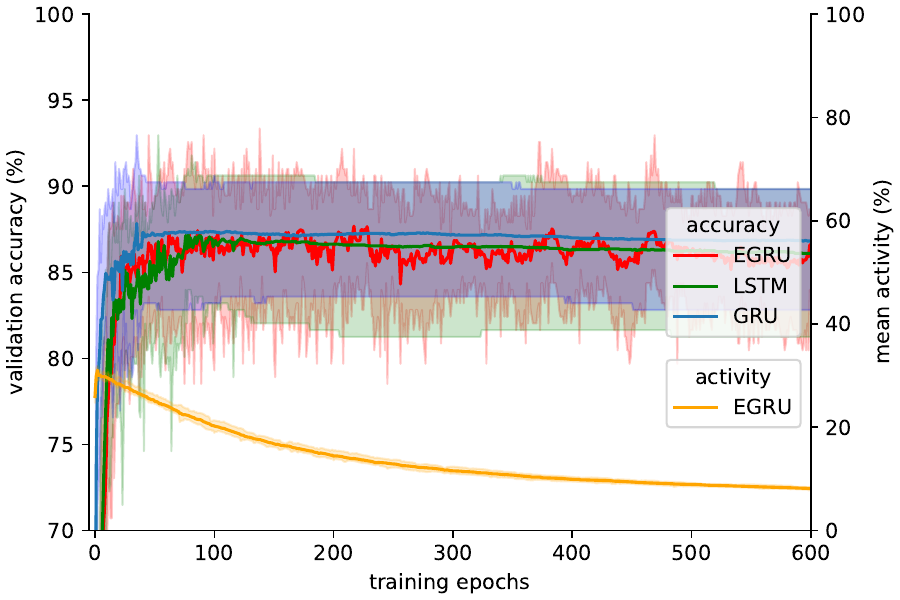}
    \caption{}
  \end{subfigure}%
  \begin{subfigure}{0.5\linewidth}
    \centering
    \includegraphics[width=0.95\linewidth]{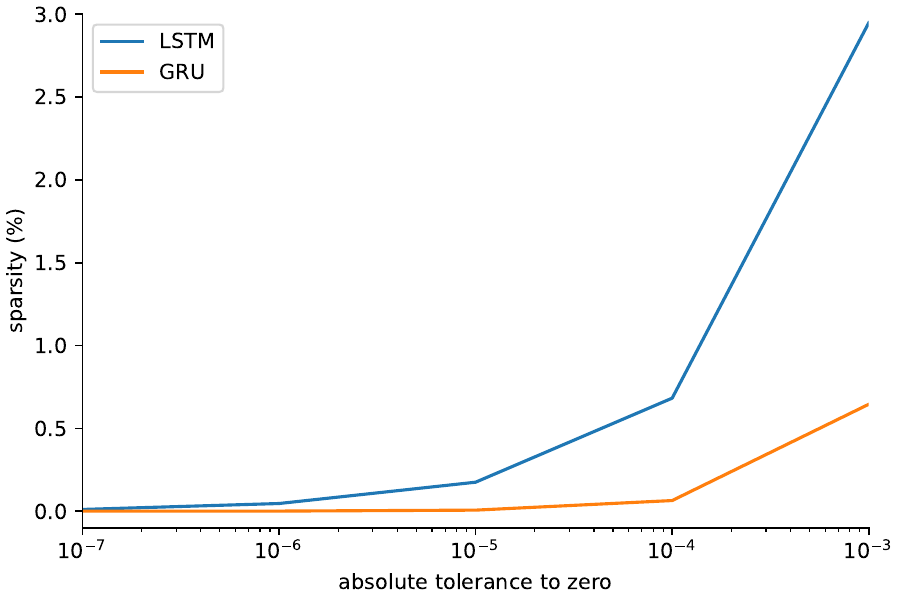}
    \caption{}
  \end{subfigure}  
  
  \caption{\textbf{(a)} mean training curves over 5 runs for DVS gesture task. \textbf{(b)} activity sparsity of LSTM and GRU for DVS gesture task over 1 run across various values of absolute tolerance to zero.}  
  \label{fig:gesture-training}
\end{figure}

\begin{figure}[p]
    \centering
    \includegraphics[width=.7\textwidth]{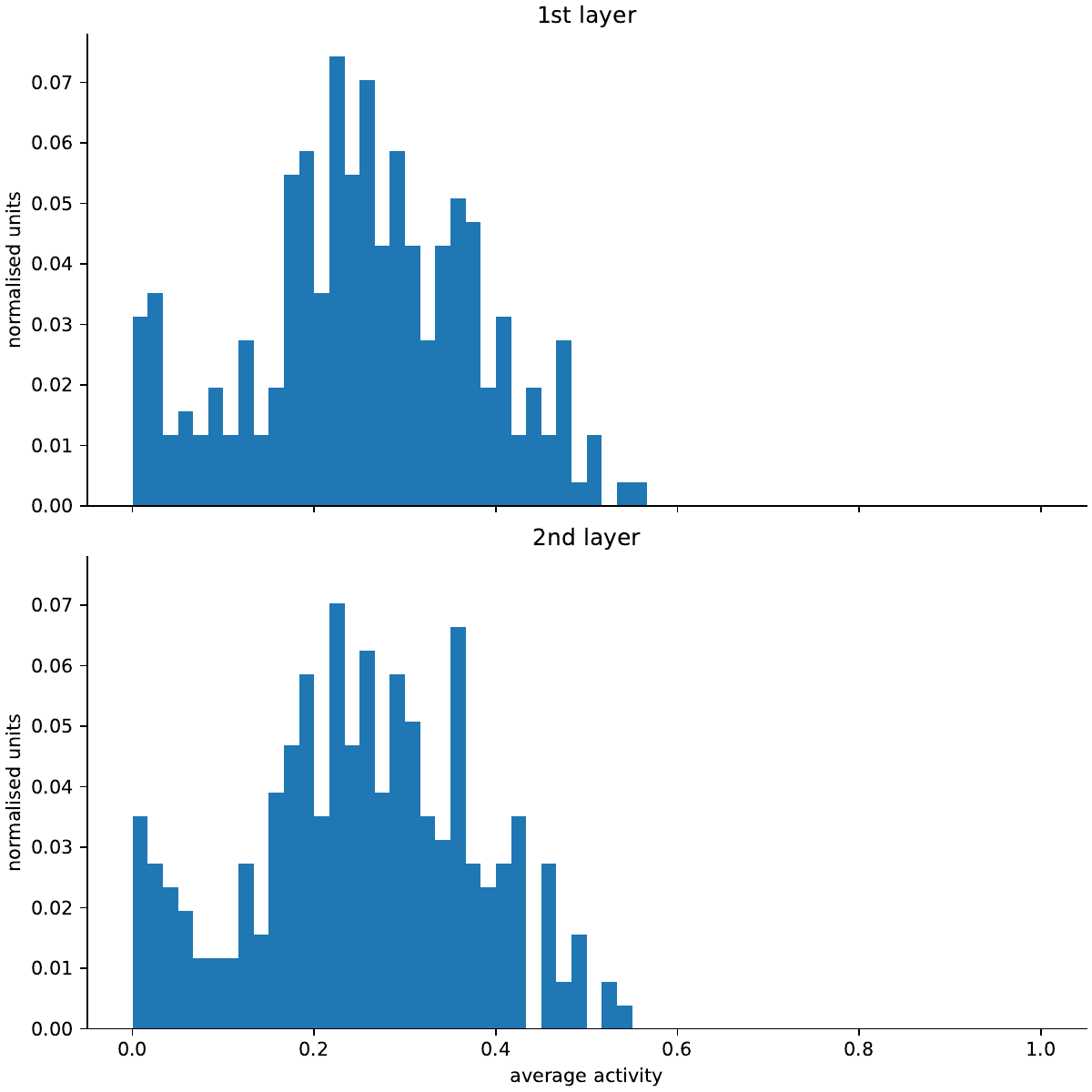}
    \caption{Histogram of normalised unit activity for fully trained 2 layer EGRU network with 256 units performing DVS gesture task. Activity is strongly skewed towards low values. There are no units always active, however some units are inactive for the entire test dataset.
    }
    \label{fig:gesture-activity-histogram}
\end{figure}

Fig. \ref{fig:gesture-training}(a) shows comparison of training curves for LSTM, GRU and EGRU,  mean activity of the EGRU network is also shown, the network achieved 80\%+ sparsity without significant drop in accuracy.
The activities of LSTM and GRU are not shown in Fig \ref{fig:gesture-training}(a) since they are always 100\%. 
In our experiments we calculate sparsity of these networks as average number of activations close to zero with an absolute tolerance of $1\times10^{-8}$, however in Fig. \ref{fig:gesture-training}(b) we show that even if we increase the absolute tolerance to $1\times10^{-3}$, the sparsity of these networks is still an order of magnitude lower than EGRU.
The analyse the activity of the individual units of EGRU network in Fig. \ref{fig:gesture-activity-histogram} with an histogram of unit activity for the entire test dataset. 
The activity of the units shown in x-axis is normalised to the sequence length. 
As expected from the overall activity sparsity shown in Table. \ref{tab:dvsgesture}, most of the units have low activity with some dead units.

Hyperparameters were chosen by conducting a grid search over the number of units (32 - 2048), number of layers (1 - 4) and values of regularization weights. Learning rate and optimizer was chosen from initial experiments. Since batch size did not have any significant effect on training, we chose a batch size that maximizes GPU utilization.
Models with CNN feature extractors are trained with slightly different hyper-parameters than the pure RNNs. These hyperparameters are chosen by a Bayesian search. This includes hidden to hidden dropout $p_h$ of 0.4 for CNN+EGRU(256) and 0.08 for CNN+EGRU(795). The batch size used in this case is 40 to ease data augmentation. The learning rate is set initially to 0.001 and then scaled by 80\% every 60 epochs. The input channels are combined into a single channel by averaging over the channel dimension.

\begin{align}
L_{reg} = w_{reg} \left( \frac{1}{N} \frac{1}{n_\text{units}} \sum_{n=1}^N \sum_{1}^{n_\text{units}}\fun{H}{ c_i\dti - \vartheta_i }\quad - 0.05 \right )\\
L_{act} = w_{v} \left( \frac{1}{N} \frac{1}{n_\text{units}} \sum_{n=1}^N \sum_{i=1}^{n_\text{units}}{c_{i} - (\vartheta_i - 0.05)} \right )
\label{eq:dvs-loss}
\end{align}
where $N$ spans mini-batch.

\begin{table}[p]
\centering
\begin{tabular}{@{}lccccc@{}}
\toprule
architecture         & para- & effective    & accu- & activity  & backward \\
(\# units)         & meters & MAC    & racy &  sparsity &  sparsity\\ 
    &           &       & (\%)  & (\%)  & (\%) at epoch 100\\
    &            &  (mean$\pm$std) & (mean$\pm$std) &  (mean$\pm$std) &  (mean$\pm$std) \\\midrule
LSTM (867)       & 16.3M     & 14.2M & 87.9$\pm$1.0       & 0    & -            \\
GRU (1024)       & 15.7M     & 10.6M & 88.1$\pm$0.8        & 0   & -             \\
GRU (1024)+DA       & 15.7M     & 10.6M &   94.8$\pm$0.3      & 0   & -             \\
\textbf{EGRU} (512)    & 5.5M     & 1.2M$\pm$0.1M & 86.0$\pm$1.2         & 76.1$\pm$5.9 & 45.7$\pm$0.7         \\
\textbf{EGRU} (1024)     & 15.7M     & 3.1M$\pm$0.4M & 87.7$\pm$2.1         & 79.8$\pm$3.3 & 54.4$\pm$1.2           \\
\textbf{EGRU} (1024+DA)     & 15.7M     & 2.7M$\pm$0.3M & 95.9$\pm$0.7         & 84.2$\pm$3.0 & 52.8$\pm$3.3           \\
\textbf{EGRU} (1024)*     & 110.1M     & 105.2M$\pm$441.3K & 85.7$\pm$0.9 & 77.3$\pm$7.0 & 64.6$\pm$1.1      \\
CNN+GRU (136)+DA*     & 1.7M+0.4M**     & 0.3M\textsuperscript{\textdagger} & 97.15$\pm$0.2 & 0 & -      \\
CNN+\textbf{EGRU} (256)+DA*     & 1.7M+1.0M**     & 0.5M$\pm$12.4,K\textsuperscript{\textdagger} & 96.8$\pm$0.3 & 73.4$\pm$2.1 & 55.3$\pm$1.3      \\
CNN+\textbf{EGRU} (795)+DA*     & 1.7M+3.1M**     & 1.6M$\pm$34.3K\textsuperscript{\textdagger} & 97.3$\pm$0.4 & 77.6$\pm$1.8 & 72.7$\pm$1.0      \\
\bottomrule
\end{tabular}
\caption{
\label{tab:suppl-dvsgesture}
Model performance over 5 runs for the DVS Gesture recognition task.  
Effective number of MAC operations as described in section \ref{sec:sparse-computation}. \textbf{*} indicates network with $128\times128$ input size, all other networks have scaled input as explained in Section \ref{sec:results-gesture}.
\textbf{**} Indicated parameters are split between CNN and RNN.
\textbf{\textdagger} Only RNN MAC operations. CNN adds an additional 79M MAC operations, however since these are not affected by activity sparsity, we exclude them from this table for brevity.
}
\end{table}

\begin{table}[p]
\centering
\begin{tabular}{@{}lll@{}}
\toprule
layer           & channels & output shape  \\ \midrule
Input           & 1        & 128x128           \\
Convolution     & 64       & 31x31             \\
ReLU            & 64       & 31x31             \\
Pooling         & 64       & 15x15             \\
Convolution     & 192      & 15x15             \\
ReLU            & 192      & 15x15             \\
Pooling         & 192      & 7x7               \\
Convolution     & 384      & 7x7               \\
ReLU            & 384      & 7x7               \\
Pooling         & 384      & 3x3               \\
Convolution     & 256      & 3x3               \\
ReLU            & 256      & 3x3               \\
Pooling         & 256      & 1x1               \\
Convolution     & 256      & 1x1               \\
ReLU            & 256      & 1x1               \\
Fully connected & 512      & 1x1               \\
ReLU            & 512      & 1x1               \\ \bottomrule
\end{tabular}
\caption{Details of CNN layers for the feature extraction head used in CNN+EGRU models}
\label{tab:dvs-cnn-layers}
\end{table}

\begin{table}[]
    \centering
    \begin{tabular}{@{}clcccc@{}}
    \toprule
        Model & EGRU & CNN+EGRU & CNN+EGRU  & \updated{CNN+GRU}     \\
        Hidden units & 512/1024 & 795 & 256 & 136         \\
    \midrule
  Layers                        & 2           & 1           & 2           & 1           \\
  Learning rate                 & \num{0.001} & \num{0.001} & \num{0.001} & \num{0.001} \\
  Learning rate decay           & 0.8         & 0.874       & 0.8         & 0.89        \\
  Learning rate decay epochs    & 100         & 56          & 100         & 70          \\
  Batch size                    & 256         & 40          & 40          & 40          \\
  Dropout $p_l$                 & 0           & 0.632       & 0           & 0.239       \\
  DropConnect $p_h$             & 0           & 0.081       & 0.4         & 0.074       \\
  Zoneout $p_z$                 & 0           & 0.2         & 0           & 0           \\
  Activity regularization       & 0.01        & 0.01        & 0.01        & -           \\
  Surrogate gradient $\epsilon$ & 1           & 0.588       & 1           & -           \\
  Threshold init $\mu$          & 0           & -0.246      & 0           & -           \\ \bottomrule
    \bottomrule
    \end{tabular}
    \caption{Detailed hyper-parameters of our best models for DVS Gesture recognition.  }
    \label{tab:dvs-params}
\end{table}

\subsubsection{Ablation study}
\label{sec:ablation-study}
We performed ablation studies, showing the performance of the EGRU models with variation of the gating mechanism.
All models in this study are a variation of our \textbf{EGRU}(1024) model.
The results of these experiments are presented in Table \ref{tab:thr-ablation}. 
By using a scalar threshold $\vartheta$ where all units share a same threshold parameter we find that the accuracy drops by 2\% but the the activity sparsity is increased to 90\%.

Next, we evaluate a model with `hard reset' where the auxiliary internal state $c_i\dti$ is set to $0$ every time an event is emitted by an unit.
We observe a drop in accuracy possibly because the hard reset loses information when the internal state has gone above threshold at at any particular simulation time step, which may happen due to the limitations on precision in discrete time simulations with a fixed time grid.
This drop in performance might be significant for applications which require high temporal resolution, which necessitates the term $-\, y_i\dtim1$ in Eq.~\eqref{eq:event-mechanism}. 
Model is also evaluated with `no reset' where the term $-\, y_i\dtim1$ is removed from Eq.~\eqref{eq:event-mechanism} which results in slightly lower accuracy and sparsity.

\begin{table}[p]
\centering
\begin{tabular}{@{}lcc@{}}
\toprule
model              & accuracy (\%) & activity sparsity (\%) \\ \midrule
Full \textbf{EGRU} (1024)& 90.2    & 82.5             \\
without regularization & 89.3 & 76.5            \\
scalar $\vartheta$ & 88.3     & 90.8              \\
hard clear         & 87.2     & 90.0                \\ 
no clear           & 88.9    & 80.7                \\
\bottomrule
\end{tabular}
\caption{Performance of the \textbf{EGRU} (1024) model for the ablation study performed on the DVS gesture task as described in Section \ref{sec:ablation-study}.}
\label{tab:thr-ablation}
\end{table}

\subsection{Sequential MNIST}
\label{sec:smnist-supplement}\label{suppl-sec:exp-details-smnist}
All the weights were initialised using Xavier uniform distribution, while the biases were initialised using a uniform distribution.
The unit thresholds were initialised using a normal distribution with mean 0 and standard deviation of $\sqrt{2}$, but was transformed to be between 0 and 1 by passing through a standard sigmoid/logistic function after every update.
We used a batch size of 500 for sMNIST and 300 for psMNIST.
In all the experiments, we trained the network with Adam with default parameters (0.001 learning rate, $\beta_1 = 0.9$, $\beta_2=0.999$) on a cross-entropy loss function.
We used gradient clipping with a max gradient norm of 0.25.
We trained models for 200 epochs for sMNIST and 700 epochs for psMNIST.
The model trained on psMNIST used DropConnect~\cite{DropConnect} with $p=0.4$.
The outputs of all the units were convolved with an exponential filter with time constant of 10 time units i.e. with $e^{\frac{-1}{10}}$ to calculate an output trace.
The value of this trace at the last time step was used to predict the class through a softmax function.

For sMNIST, hyper-parameters were chosen by performing a search over batch sizes (50-1000), learning rates ($10^-3, 10^-4$), use of output trace, activity regularisation. 
An extensive Bayesian search was conducted using Weights \& Biases~\citep{wandb} to optimize hyperparameters of EGRU with 590 hidden units on psMNIST on NVIDIA V100 GPUs.
The initialisation method of the thresholds were also tweaked -- currently we use a normal initialisation with a sigmoid projection into the $[0,1]$ range, but we experimented with projecting it with an absolute value followed by clipping, which proved unstable.

\begin{table}[p]
    \centering
    \begin{tabular}{@{}l|c|c|c|c|c}
        \toprule
        architecture & parameters & effective  & test     & activity  & backwards \\
                  (\# units)   &            &  MAC       & accuracy &  sparsity & sparsity (\%) at \\
                               &            &            &   (\%)       &    (\%)       & epochs 20/50/100 \\
                               &            &  (mean$\pm$std) & (mean$\pm$std) &  (mean$\pm$std) &  (mean$\pm$std) \\
        \midrule
        GRU (512) & 791K & 795K & 98.6$\pm$0.2 & - & - \\
        GRU (590) & 1.049M & 1.054M & 98.7$\pm$0.1 & - & - \\
        \textbf{EGRU} (512) & 790K & (147$\pm$7)K & 87.2$\pm$3.0 & 82.1$\pm$0.9 & 22.2$\pm$2.8/24.9$\pm$0.7/  \\
        & & & & & 28.7$\pm$ 1.1 \\
        \textbf{EGRU} (590) & 1.048M & (210$\pm$51)K & 95.5$\pm$1.6 & 80.5$\pm$4.9 & 24.9$\pm$6.8 / 26.1$\pm$5.9 /  \\
        & & & & & 25.6$\pm$1.7 \\
        \bottomrule
    \end{tabular}
\caption{
\label{tab:suppl-smnist}
Model performance over 4 runs for sequential MNIST task. 
Test scores are given as percentage accuracy, where higher is better.
}

\end{table}

\subsection{PTB Language modeling}
\label{suppl-sec:exp-details-ptb}
Our experimental setup largely follows \cite{merity_Regularizing_2017}.
In particular, we download and preprocess PennTreebank~\citep{PennTreebank} and WikiText-2~\citep{WikiText} with their published code~\footnote{\url{https://github.com/salesforce/awd-lstm-lm}}.
Words are projected to an $d_{\text{emb}}$-dimensional dense vector by a linear transformation, followed by three RNN layers without skip connections.
The first two RNN layers feature the same hidden dimension, while the hidden dimension of the last RNN layer equals the word vector embedding dimension.
As common in language modeling, we apply cross entropy loss and use weight tying \cite{WeightTyingInan, WeightTyingPress}.

\subsubsection{Training details and hyperparameter optimization}
\label{suppl-subsec:training-ptb}
\begin{figure}
    \centering
    \includegraphics[width=0.9\textwidth]{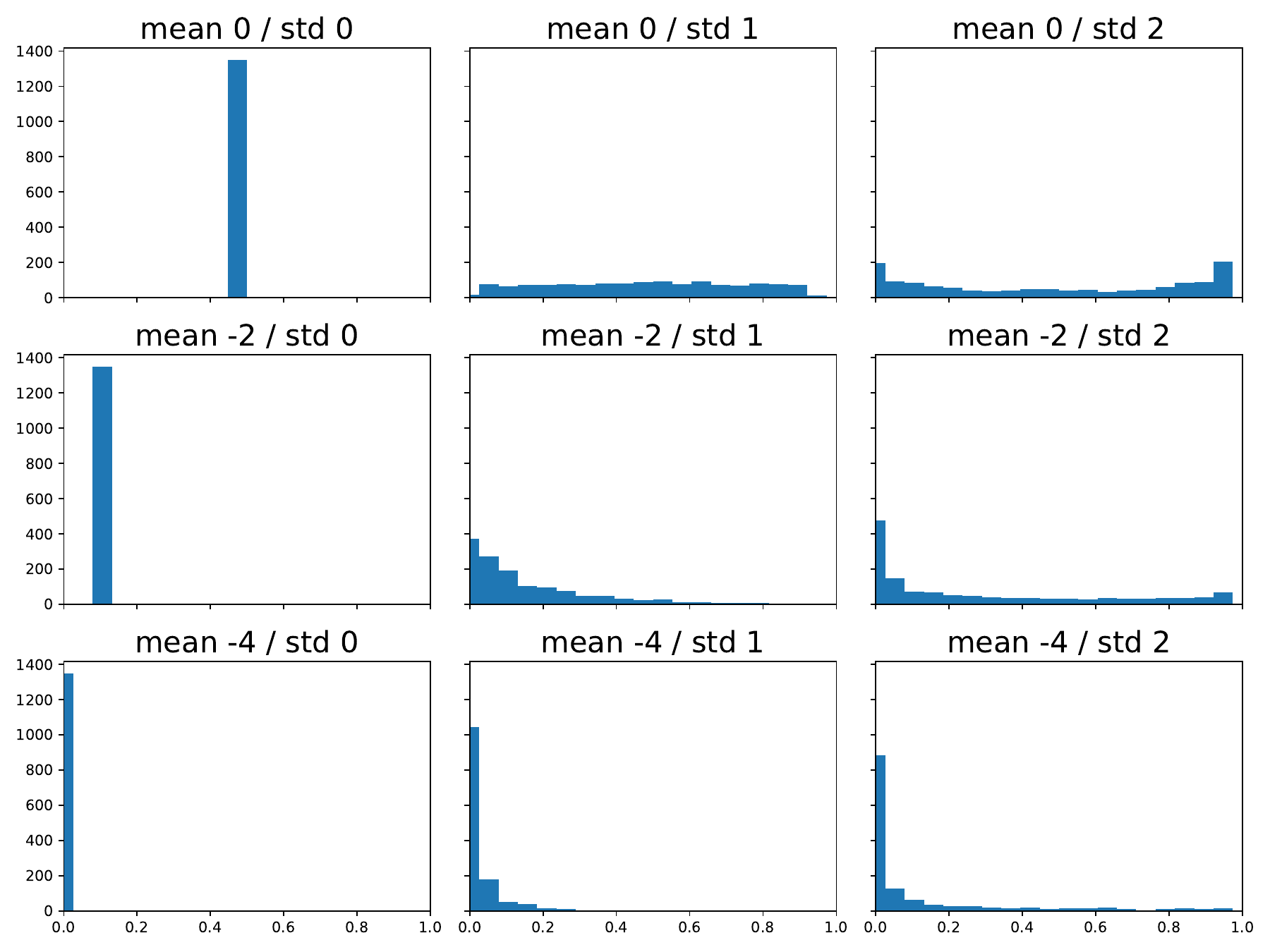}
    \caption{Visualization of threshold initializations for different parameters $\mu$ and $\sigma$.
    The thresholds are reparameterized with a sigmoid function to limit their domain to the interval $[0, 1]$.
    With $\tau_i$ drawn from a normal distribution $\tau_i \sim \mathcal{N}(\mu, \sigma\sqrt{2})$,
    the thresholds are initialized as $\varphi_i = 1 / (1 + \exp(-\tau_i))$, where $\tau_i$ are the trainable parameters}
    \label{fig:threshold-initialization}
\end{figure}

Our activity sparsity mechanism introduces two new hyperparameters $\epsilon$ and $\mu$.
First, the shape of the surrogate gradient 
\begin{equation}
    \frac{\mathrm{d}H}{\mathrm{d}c} = \lambda ~ \max
    \left(1 - \lvert c \rvert / \epsilon\right)
\end{equation}
is defined by $\epsilon$, which thus determines the backward sparsity.
Second, the initialization of the rectifyer thresholds $\varphi$ determines both inference and BPTT sparsity at initialization.
We choose to reparameterize thresholds with a sigmoid function to limit their domain to the interval $[0, 1]$.
With $\tau_i$ drawn from a normal distribution $\tau_i \sim \mathcal{N}(\mu, \sigma\sqrt{2})$,
the thresholds are initialized as $\varphi_i = 1 / (1 + \exp(-\tau_i))$, where $\tau_i$ are the trainable parameters and $\mu$ is the new hyperparameter.
See Figure \ref{fig:threshold-initialization} for the resulting distribution of initial thresholds.
Table \ref{tab:thresholds} shows the sensitivity of model performance w.r.t. these parameters.
Non-trainable thresholds are also considered in Table \ref{tab:thresholds}.
We observe that language modeling benefits from initialization near 0.
Trainable thresholds slightly outperform non-trainable thresholds.
The gap depends on the initialization and is fairly small for the best initialization strategies.
This is not very much surprising as the model is able to counteract constant thresholds with bias terms in the GRU equations.

We apply most of the regularization strategies of \cite{merity_Regularizing_2017}, except for (temporal) activity regularization.
Backpropagation through time is conducted with a variable sequence length.
With 95\% probability, the sequence length is drawn from $\mathcal{N}(s, 5)$, and with 5\% probability the sequence length is drawn from $\mathcal{N}(s/2, 5)$, 
where $s$ is a tuned hyperparameter.
We apply variational dropout \cite{VariationalDropout} to the vocabulary with probability $p_{\text{voc}}$, to the word embedding vectors with probability $p_{\text{emb}}$ as well as to each layer output with probability $p_l$.
DropConnect \cite{DropConnect} was applied to the hidden-to-hidden weight matricies with probability $p_h$.
We experimented with both Adam \cite{Adam2015} and NT-AvSGD \cite{merity_Regularizing_2017} optimization procedures.
While Adam lead to competitive results for all models, GRU based models did not converge to competitive results using NT-AvSGD. 
When optimized with SGD based optimizers, both GRU and EGRU fell behind Adam optimized models.
Momentum was set to 0 as reported in \cite{Melis2018}.
Gradient clipping was applied to all models, where the magnitude of clipped gradients only made very small differences in results.
While gradient clipping of 0.25 was used for GRU, we used 2.0 for EGRU.

We apply a cosine-annealing learning rate schedule, where the first $n/2$ epochs were trained at constant learning rate $\lambda$, 
and a cosine decay from $\lambda$ to $0.1\cdot\lambda$ was applied for the remaining $n/2$ epochs.
All EGRU models were trained for 2500 epochs.

An extensive Bayesian search was conducted using Weights \& Biases~\citep{wandb} to optimize hyperparameters of GRU with 1350 hidden units and EGRU with 1350 and 2000 hidden units on Penn Treebank for about 65~GPU days on NVIDIA A100 GPUs.
The surrogate gradient parameter $\epsilon$ and the initialization of the thresholds $\varphi_i$ are treated as hyperparameters of this model.
Due to our constrained computational resources, we used the same hyperparameters on WikiText-2.

We found the word embedding dimension $d_{\text{emb}} = 400$ set by \citep{merity_Regularizing_2017} to be a good fit for GRU.
For EGRU, we observed much larger dimensions around $d_{\text{emb}} = 800$ to outperform smaller dimensions.
This increases the number of parameters of the word-embedding layer by about a factor of 2.
Language models need to compare the output embedding vector via dot product with the embedding vectors of the dictionary. 
Since EGRU outputs only positive values, we hypothesise that extra parameters are required to cancel terms in the dot-product.
See table \ref{tab:ptb-params} for detailed hyperparameters of the best models.

\begin{table}[]
    \centering
    \begin{tabular}{@{}c l|c|c|c@{}}
    \toprule
        &Model & GRU & \textbf{EGRU} & \textbf{EGRU} \\
        &Hidden units & 1350 & 1350 & 2000 \\
    \midrule
        \multirow{5}{*}{PTB} & Test ppl (best) & \num{66.3} & \num{58.7} & \num{58.8} \\
        &Val perplexity (best) & 68.7 & 59.5 & 59.6 \\
        &Val perplexity (mean~$\pm$~std) & \num{68.9 \pm 0.1} & \num{59.7 \pm 0.1} & \num{60.0 \pm 0.5} \\
        &Forward sparsity (test) & \SI{0.0 \pm 0.0}{\percent} & \SI{79.9 \pm 0.1}{\percent} & \SI{85.3 \pm 0.9}{\percent} \\
        &Backward sparsity (train) & \SI{0.0 \pm 0.0}{\percent} & \SI{46.0 \pm 0.3}{\percent} & \SI{40.5 \pm 3.8}{\percent} \\
        &Effective MACs (RNN~+~emb) & (21.9~+~5.6)\si{\million} & (6.8~+~3.1)\si{\million} & (9.7~+~3.0)\si{\million} \\
    \midrule
        \multirow{5}{*}{WT2} & Test perplexity (best) & 71.8 & 70.6 & 68.9 \\
        &Val perplexity (best) & 75.7 & 73.9 & 71.5 \\
        &Val perplexity (mean~$\pm$~std) & \num{75.9 \pm 0.1} & \num{74.0 \pm 0.1} & \num{75.7 \pm 6.5} \\
        &Forward sparsity (test) & \SI{0.0 \pm 0.0}{\percent} & \SI{77.0 \pm 0.1}{\percent} & \SI{84.6 \pm 2.9}{\percent} \\
        &Backward sparsity (train) & \SI{0.0 \pm 0.0}{\percent} & \SI{43.8 \pm 0.3}{\percent} & \SI{36.1 \pm 8.7}{\percent} \\
        &Effective MACs (RNN~+~emb) & (21.9~+~16.9)\si{\million} & (7.4~+~10.1)\si{\million} & (10.6~+~9.5)\si{\million} \\
    \midrule
        &Learning rate & \num{4.62E-04} & \num{4.44E-04} & \num{4.94E-04} \\
        &Batch size & 96 & 64 & 128 \\
        &Sequence length $s$ & 34 & 68 & 67 \\
        &Embedding dimension $d_{\text{emb}}$ & 563 & 788 & 786 \\
        &Dropout $p_h$ & 0.506 & 0.679 & 0.621 \\
        &Dropout $p_l$ & 0.474 & 0.264 & 0.241 \\
        &Dropout $p_{\text{emb}}$ & 0.729 & 0.707 & 0.765 \\
        &Dropout $p_{\text{voc}}$ & 0.093 & 0.055 & 0.149 \\
        &Weight decay & \num{4.60E-06} & \num{9.01E-06} & \num{6.69E-06} \\
        &Activity regularization & 2.766 & 0 & 0 \\
        &Temporal Activity regularization & 0.29 & 0 & 0 \\
        &Surrogate gradient $\epsilon$ & - & 0.459 & 0.425 \\
        &Threshold init $\mu$ & - & -3.769 & -3.496 \\
    \bottomrule
    \end{tabular}
    \caption{Detailed results and parameters for our best models. 
    Mean and standard deviations are calculated over 5 runs with different random seeds.
    Effective MAC operations consider the layer-wise sparsity in the forward pass. 
    Activity sparsity is given for the trained model to resemble inference sparsity. 
    Backward sparsity is averaged over the whole training.
    Model parameters were optimized on Penn Treebank and transfered to WikiText-2. }
    \label{tab:ptb-params}
    \label{tab:suppl-ptb}
\end{table}

\subsection{Model compression through activity pruning}
\label{suppl-subsec:pruning-ptb}
\begin{figure}
    \centering
    \includegraphics[width=\textwidth]{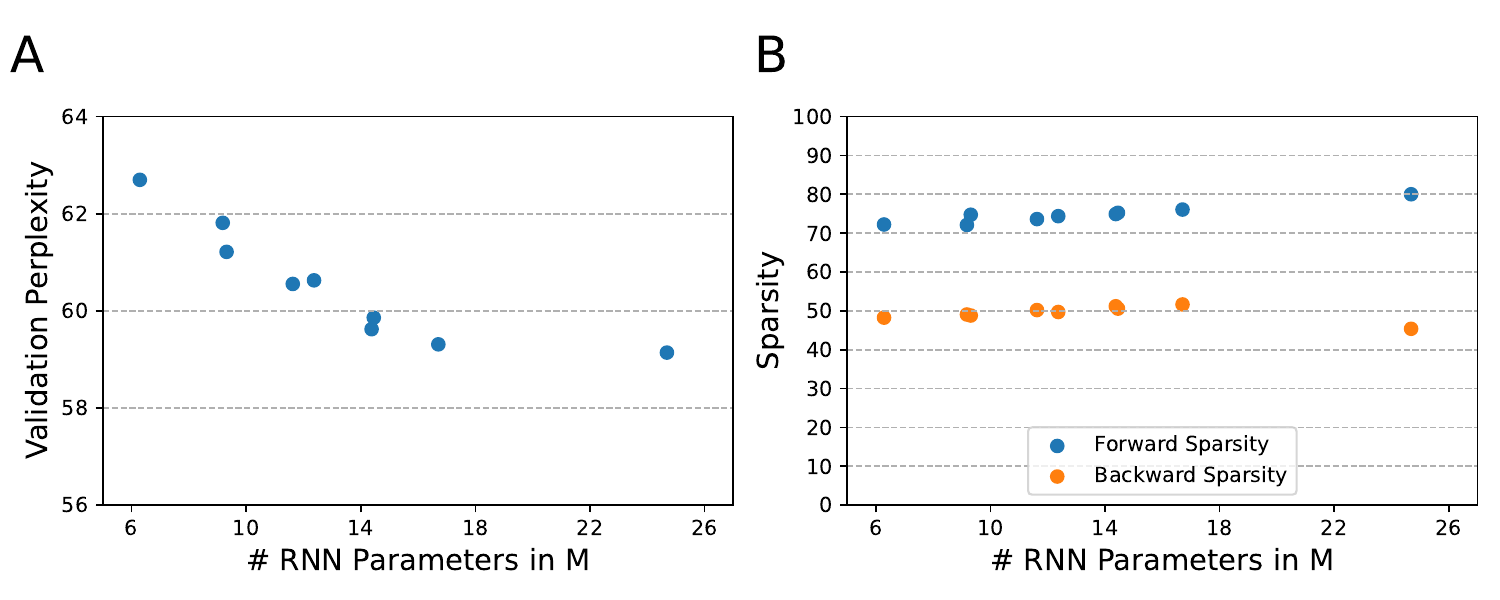}
    \caption{We apply different levels of layer-wise model compression according to Sec. \ref{suppl-subsec:pruning-ptb}.
    \textbf{A} Model performance 
    \textbf{B} Forward and backward sparsity }
    \label{fig:ptb-pruning}
\end{figure}

We present a simple model compression heuristic based on activity.
Starting from a trained model, we remove the least active $r_i$~\% of the units of layer $i$.
Since we observed different levels of activity in the layers,
we work with different combinations of compression rates $r_i$ per layer.
Figure \ref{fig:ptb-pruning} shows how model performance and sparsity depend on the model compression.
Surprisingly, we observe very similar sparsity levels across the evaluated compressed models.

\section{Dataset licenses}
\label{suppl-sec:dataset-licenses}
Penn Treebank \cite{PennTreebank} is subject to the Linguistic Data Consortium User Agreement for Non-Members\footnote{\url{https://www.ldc.upenn.edu/data-management/using/licensing}}.
\begin{quote}
LDC Not-For-Profit members, government members and nonmember licensees may use LDC data for noncommercial linguistic research and education only. For-profit organizations who are or were LDC members may conduct commercial technology development with LDC data received when the organization was an LDC for-profit member unless use of that data is otherwise restricted by a corpus-specific license agreement. Not-for-profit members, government members and nonmembers, including nonmember for-profit organizations, cannot use LDC data to develop or test products for commercialization, nor can they use LDC data in any commercial product or for any commercial purpose.
\end{quote}
Following \cite{merity_Regularizing_2017}, we download Penn Treebank data from \url{http://www.fit.vutbr.cz/~imikolov/rnnlm/simple-examples.tgz}.

The DVS128 Gesture Dataset \citep{amir2017low} is released under the Creative Commons Attribution 4.0 license and can be retrieved from: \url{https://research.ibm.com/interactive/dvsgesture/}.
We used Tonic library \citep{tonic_pytorch} for Pytorch to preprocess data and to apply transformations.

The sequential MNIST task~\citep{le2015simple} is based on the MNIST dataset first introduced in~\citep{lecun1998gradient}, available from: \url{http://yann.lecun.com/exdb/mnist/}.
\section{Hardware and software details}
\label{suppl-sec:hardware-software}
We implement EGRU as a modification of Haste GRU~\citep{haste2020} and observe slightly shorter wallclock times than PyTorch's~\citep{pytorch} GRU implementation.

Most of our experiments were run on NVIDIA A100 GPUs.
Some initial hyper-parameter searches were conducted on NVIDIA V100 and Quadro RTX 5000 GPUs.
We used about 45,000 computational hours in total for training and hyper-parameter searches.
All models and experiments were implemented in PyTorch.
For the continuous time EGRU model, we also used the torchdiffeq~\citep{chen_Neural_2018} library.

The machines used for the DVS128 gesture recognition task and for the PTB language modeling task feature 8x NVIDIA A100-SXM4 (40GB) GPUs, 2x AMD EPYC CPUs 7352 with 24 cores each, and 1TB RAM on each compute node.
For each run, we only use a single GPU, and a fraction of the cores and memory available on the node to run multiple experiments in parallel.
The nodes operate Red Hat Enterprise Linux Server (release 7.9).

\clearpage

\begin{figure}[p]
    \centering
    \includegraphics[width=.6\textwidth]{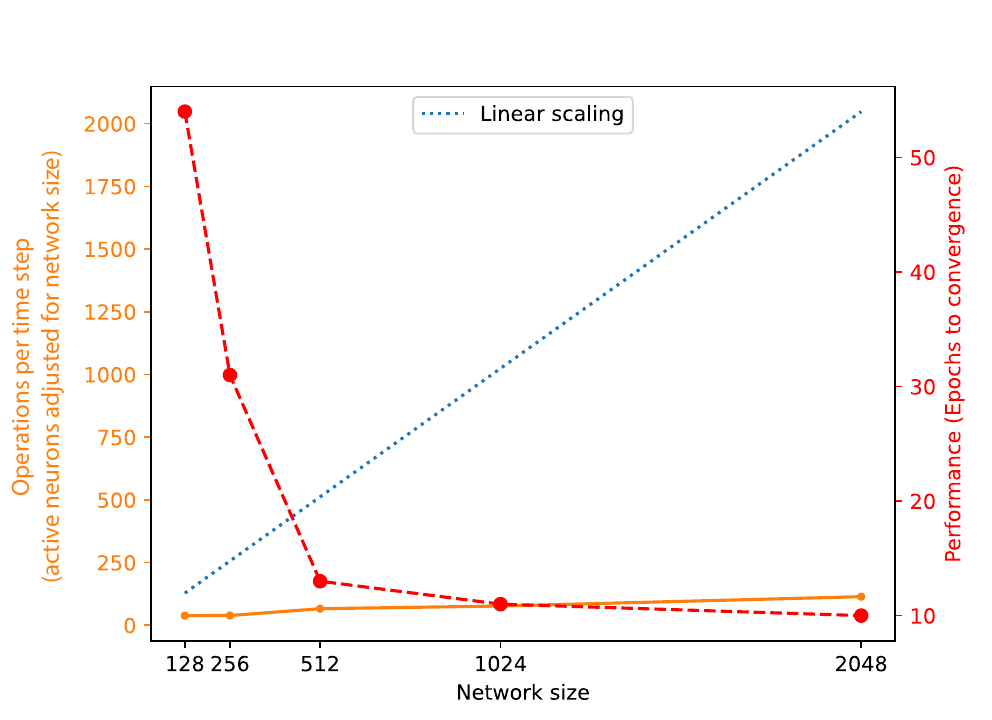}
    \caption{Illustration of the scaling properties of the EGRU on a $14\times14$ sequential MNIST task (1 run per network size).
    As the size of the network increases, the network converges faster.
    Increasing the network size 10x increases the speed of convergence 5x, while increasing the total amount of computation per sample only 2x.
    The total amount of computation is adjusted for network size.
    The smaller subsampled 14x14 sMNIST task was chosen here for reasons of computational limitations.
    }
\end{figure}

\begin{figure}
    \centering
    \begin{subfigure}{0.48\linewidth}
        \includegraphics[width=\linewidth]{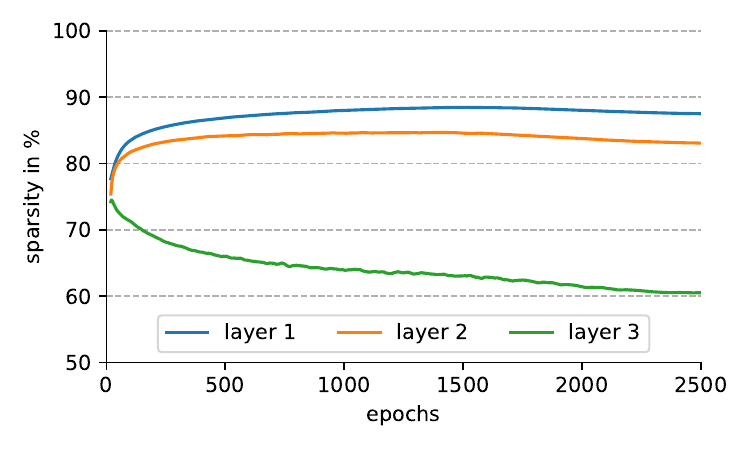}
        \caption{}
    \end{subfigure}
    \hfill
    \begin{subfigure}{0.48\linewidth}
        \includegraphics[width=\linewidth]{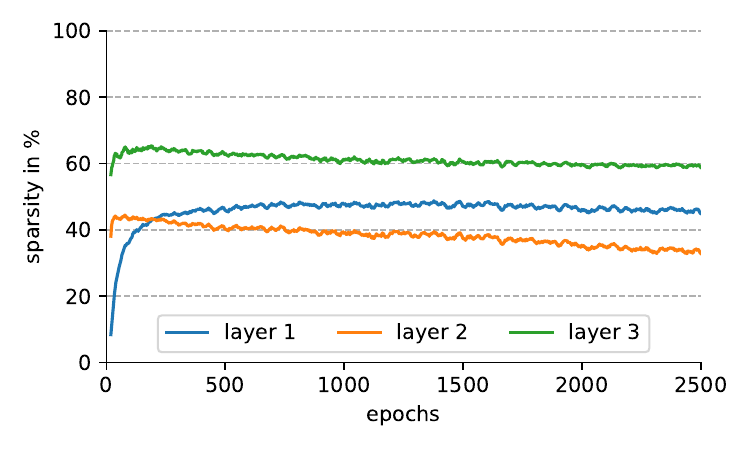}
        \caption{}
    \end{subfigure}
    \caption{EGRU with 1350 hidden units on the Penn Treebank language modeling task with pseudo-derivative $\epsilon = 0.45$ . \textbf{(a)} layer-wise forward sparsity \textbf{(b)} layer-wise backward sparsity}
    \label{fig:ptb-sparsity}
\end{figure}

\begin{figure}
    \centering
    \includegraphics[width=0.8\textwidth]{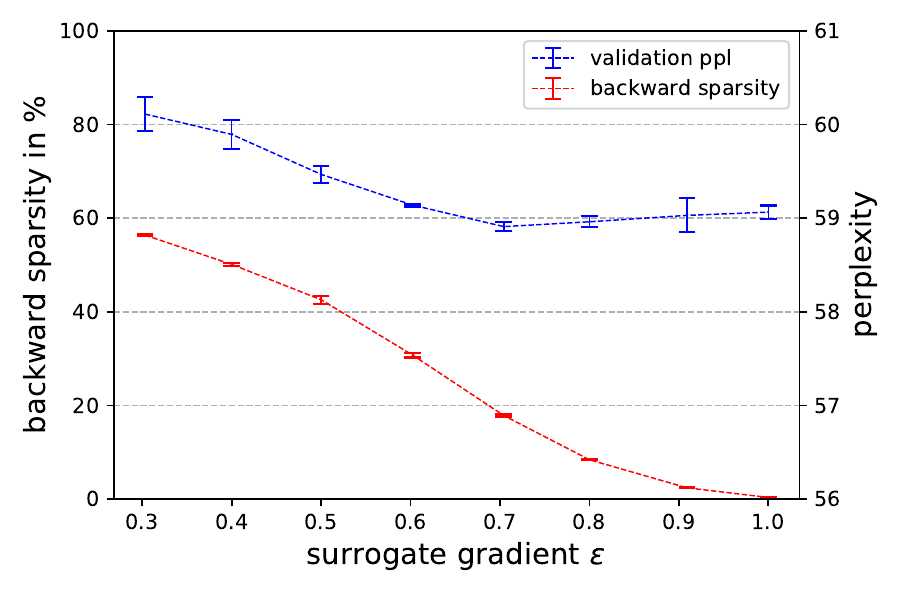}
    \caption{Backward sparsity and corresponding perplexity for a 3-layer EGRU with 1350 hidden units on the Penn Treebank language modeling task with varying pseudo-derivative support $\epsilon$.
    Standard deviations are calculated over three runs with different random seeds.}
    \label{fig:ptb-epsilon}
\end{figure}

\clearpage
\updated{
\begin{table}[]
    \centering
    \begin{tabular}{cc|ccc}
    \toprule
\multicolumn{2}{c}{\color{gray} non-trainable thresholds} & \multicolumn{3}{c}{threshold initialization std} \\
\multicolumn{2}{c}{trainable thresholds} & 0.0 & 1.0 & 2.0\\
\midrule
\multirow{6}{*}{threshold initialization mean}  & \multirow{2}{*}{-4.0} & {\color{gray} \num{59.7 +- 0.1}} & {\color{gray} \num{59.5 +- 0.1}} & {\color{gray} \num{60.1 +- 0.1}} \\
  &  & \num{59.2 +- 0.1} & \num{59.4 +- 0.2} & \num{60.0 +- 0.2} \\
\cmidrule{3-5}
  & \multirow{2}{*}{-2.0} & {\color{gray} \num{682.7 +- 0.0}} & {\color{gray} \num{63.0 +- 0.2}} & {\color{gray} \num{63.9 +- 0.2}} \\
  &  & \num{682.7 +- 0.0} & \num{61.8 +- 0.2} & \num{62.6 +- 0.2} \\
\cmidrule{3-5}
  & \multirow{2}{*}{0.0} & {\color{gray} \num{682.7 +- 0.0}} & {\color{gray} \num{292.5 +- 337.9}} & {\color{gray} \num{73.8 +- 0.6}} \\
  &  & \num{682.7 +- 0.0} & \num{288.8 +- 341.1} & \num{69.8 +- 0.1} \\
    \bottomrule
    \end{tabular}
    \caption{
    Performance on Penn Treebank word-level language modeling for different values of the initialization parameters $\mu$ and $\sigma$ (see \ref{suppl-sec:exp-details-ptb}).
    See table \ref{fig:threshold-initialization} for the corresponding distribution of thresholds
    Scores are given as perplexity averaged over three runs, where lower is better.
    Gray values correspond to models with non-trainable thresholds.}
    \label{tab:thresholds}
\end{table}
}
\bibliography{references}

\makeatletter\@input{xx.tex}\makeatother